\documentclass{article}

\usepackage{arxiv}
\usepackage[symbol]{footmisc}

\usepackage[utf8]{inputenc} 
\usepackage[T1]{fontenc}    
\usepackage{hyperref}       
\usepackage{url}            
\usepackage{booktabs}       
\usepackage{amsfonts}       
\usepackage{nicefrac}       
\usepackage{microtype}      
\usepackage{lipsum}		
\usepackage{graphicx}
\usepackage{array}
\usepackage[labelfont=bf]{caption}
\usepackage[numbers,compress]{natbib}
\usepackage{doi}
\usepackage{enumitem}
\usepackage[listings]{tcolorbox}
\usepackage{listings}
\usepackage{framed}
\newcommand{\eg}{\emph{e.g., }}

\newcommand{\cf}{\emph{cf. }}

\title{Intelligent System for Automated Molecular Patent Infringement Assessment}

\date{} 					

\author{
Yaorui Shi$^{1,2,*}$,
Sihang Li$^{1,2,*}$,
Taiyan Zhang$^{1}$,
Xi Fang$^{1}$,
Jiankun Wang$^{1}$,
\\ \textbf{
Zhiyuan Liu$^{3}$,
Guojiang Zhao$^{1}$,
Zhengdan Zhu$^{1}$,
Zhifeng Gao$^{1}$,
Renxin Zhong$^{4}$,
}
\\ \textbf{
Linfeng Zhang$^{1,5}$,
Guolin Ke$^{1}$,
Weinan E$^{2,5,6,7}$,
Hengxing Cai$^{1 \dag}$,
Xiang Wang$^{2 \dag}$
}
\\ \\
$^1$ DP Technology, Beijing, China \\
$^2$ University of Science and Technology of China, Hefei, Anhui, China \\
$^3$ National University of Singapore, Singapore \\
$^4$ School of Intelligent Systems Engineering, Sun Yat-Sen University, Shenzhen, China \\
$^5$ AI for Science Institute, Beijing, China \\
$^6$ School of Mathematical Sciences, Peking University, Beijing, China \\
$^7$ Center for Machine Learning Research, Peking University, Beijing, China \\
}

\renewcommand{\undertitle}{}

\begin{document}
\maketitle

\begin{abstract}
Automated drug discovery offers significant potential for accelerating the development of novel therapeutics by substituting labor-intensive human workflows with machine-driven processes. 
However, molecules generated by artificial intelligence may unintentionally infringe on existing patents, posing legal and financial risks that impede the full automation of drug discovery pipelines.
This paper introduces PatentFinder, a novel multi-agent and tool-enhanced intelligence system that can accurately and comprehensively evaluate small molecules for patent infringement. 
PatentFinder features five specialized agents that collaboratively analyze patent claims and molecular structures with heuristic and model-based tools, generating interpretable infringement reports.
To support systematic evaluation, we curate MolPatent-240, a benchmark dataset tailored for patent infringement assessment algorithms.
On this benchmark, PatentFinder outperforms baseline methods that rely solely on large language models or specialized chemical tools, achieving a 13.8\% improvement in F1-score and a 12\% increase in accuracy.
Additionally, PatentFinder autonomously generates detailed and interpretable patent infringement reports, showcasing enhanced accuracy and improved interpretability.
The high accuracy and interpretability of PatentFinder make it a valuable and reliable tool for automating patent infringement assessments, offering a practical solution for integrating patent protection analysis into the drug discovery pipeline.
\end{abstract}

\section{Introduction}
\footnotetext[1]{Equal contribution.}
\footnotetext[2]{Correspondence authors: X.W. (\url{xiangwang1223@gmail.com}) and H.C. (\url{caihengxing@dp.tech}).}
\begin{figure}[t]
    \centering
    \includegraphics[width=\linewidth]{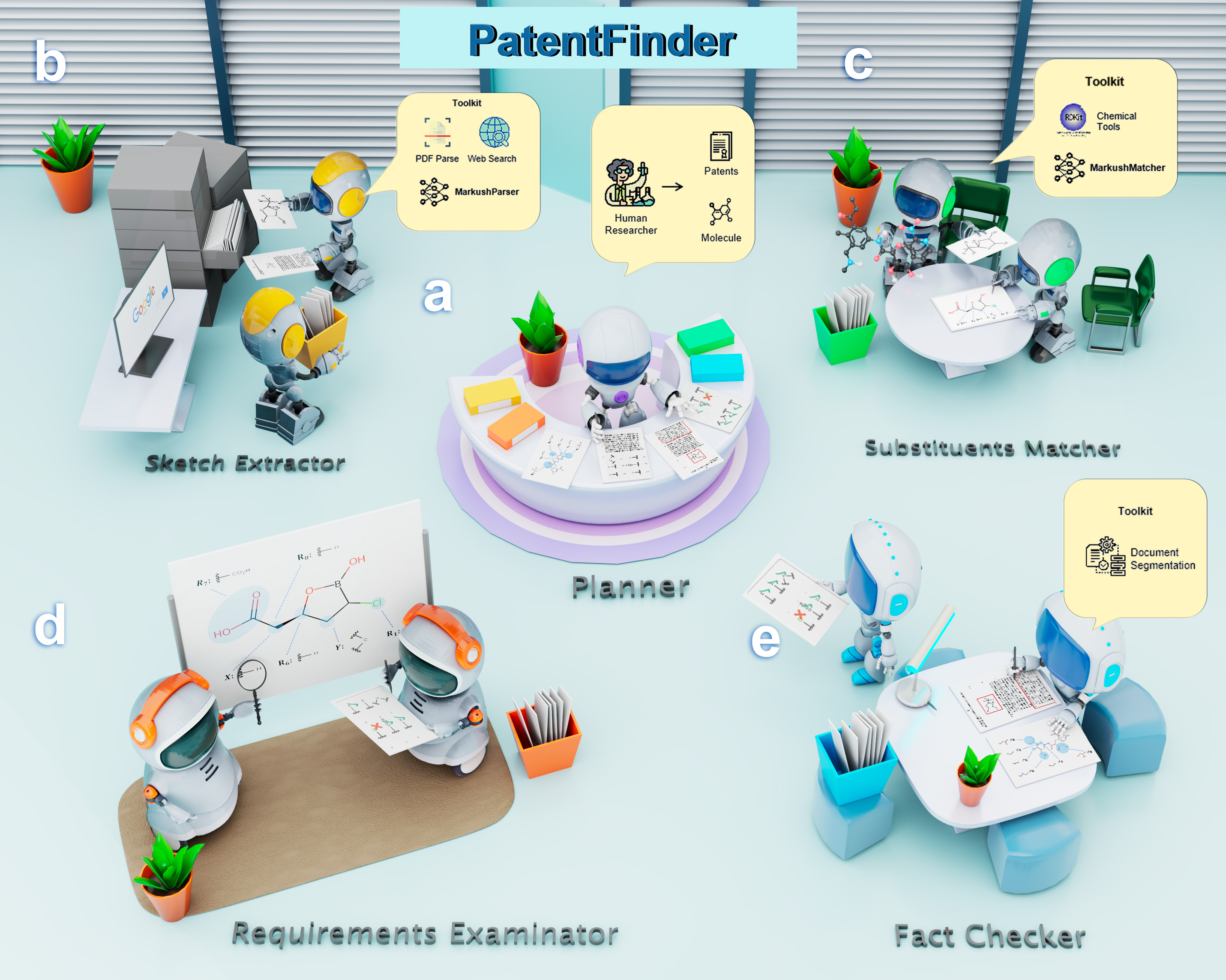}
    \caption{\textbf{Overview of PatentFinder.} Patent Finder is a multi-agent framework for autonomous molecular patent infringement assessment. a) Planner coordinates subtasks among agents and compiles a comprehensive infringement report. b) Sketch Extractor extracts the patent's core Markush structures and associated claim requirements. c) Substituents Matcher identifies and validates substituent groups in Markush expressions relative to the query molecule. d) Requirements Examinator assesses whether the query molecule meets the patent’s substituent group requirements. e) Fact Checker verifies agents’ outputs against original claims and corrects discrepancies for accuracy.}
    \label{fig:framework}
\end{figure}

Drug discovery~\cite{dd-1,dd-2} remains a cornerstone in humanity’s ongoing fight against disease, representing a vital pathway to improving global health outcomes.
Traditional methods~\cite{traditional-dd} rely on wet-lab experiments and human expertise, making the process resource-intensive and time-consuming.
Recent advances in artificial intelligence (AI) have been revolutionizing this field with automated drug discovery pipelines. AI-driven solutions have achieved remarkable successes across key tasks, including retrosynthesis planning~\cite{seq2seq, AT, r-smiles, megan, localretro}, wet lab experiment automation~\cite{coscientist, chemcrow, reactxt, smiles2actions, paragraph2actions}, molecular docking~\cite{alphafold3, diffdock, docking-1, docking-2, docking-3}, and de novo drug design~\cite{gen-dd-1,gen-dd-2,resgen, dl_for_molgen_survey, 3dmolm}. These solutions significantly reduce human intervention while enhancing the throughput and precision of drug development workflows.
However, the rapid adoption of AI in drug discovery introduces new challenges. While the AI-generated molecules may be novel, they could still fall within the scope of existing patents without intention~\cite{ChemTS}. This raises substantial legal and financial risks for pharmaceutical companies, underscoring the critical need for an integrated patent infringement assessment within AI-guided drug discovery pipelines. Current patentability assessment workflows~\cite{patent-assessment-1,patent-assessment-2}, which still rely on manual inspections by pharmaceutical chemistry experts, pose a significant bottleneck to achieving fully automated drug discovery. An intelligent, end-to-end system is demanded to streamline and enhance the patent infringement assessment process.

Large language models (LLMs)~\cite{bert, fm_opportunities, gpt3, gpt4} have demonstrated exceptional capabilities in analysis, reasoning, and problem-solving across diverse tasks~\cite{gpt3,instruct-gpt, Galactica, llama}. 
In particular, multimodal LLMs extend these capabilities to interpret images~\cite{gpt4o, llava, nextgpt, gpt4} and PDF documents~\cite{gemini}, making them increasingly viable for scientific domains like drug discovery.
These advancements also highlight the potential for integrating LLMs into patent protection analysis, enabling tasks like extracting critical information from patents and comparing the structures of new molecules against patent claims.
However, applying LLMs directly to patent infringement assessment reveals several significant obstacles:
(1) \textbf{Opaque Reasoning Paths:} 
LLMs often yield conservative evaluations that lack reliability and interpretability, making their reasoning pathways difficult to trust in high-stakes applications, like patent infringement assessment.
(2) \textbf{Inaccurate Parsing of Molecular Structures:} 
The built-in image and PDF parsers in LLMs struggle to interpret molecular diagrams and Markush notations in patents accurately.
This limitation, likely resulting from insufficient training on molecular data, especially on complex Markush structures, impairs their capability to perform fine-grained molecular analyses that patent claims require.
(3) \textbf{Lack of Benchmarks for Patent Infringement Assessment:} 
Currently, no open-source benchmark is available for patent infringement assessment, although there are some datasets designed for other patent-relevant tasks~\cite{surechembl, patcid}. The lack of benchmarks limits the systematical evaluation of relevant algorithms.

To address these challenges, we introduce PatentFinder, an innovative end-to-end intelligent system designed to accurately and comprehensively assess whether a molecule falls within the scope of a given patent. 
By leveraging a multi-agent, tool-enhanced framework, PatentFinder decomposes the inherently complex task of patent assessment into specialized, manageable subtasks, each handled by tailored agents with specialized tools.
As illustrated in Figure\ref{fig:framework}, PatentFinder consists of five core LLM agents: \textbf{Planner} -- assigns subtasks to specialized agents and compiles their outputs into a comprehensive infringement report, \textbf{Sketch Extractor} -- identifies key molecular structures and converts them into Markush expressions, \textbf{Substituents Matcher} -- verifies the substituent groups within Markush expressions relative to the target molecule, \textbf{Requirements Examinator} -- evaluates if the query molecule meets the patent’s structural requirements, and \textbf{Fact Checker} -- ensures accuracy by cross-verifying prior agents’ results against original claims and performing rollbacks if necessary. 
These agents independently engage with neural network models and heuristic tools to tackle specific subtasks, ultimately synthesizing a comprehensive patent infringement report.

The major innovations of PatentFinder include: 
(1) \textbf{Multi-Agent Framework for Interpretability:} To provide a clear reasoning path for LLMs, the multi-agent architecture enables PatentFinder to produce intermediate outputs, which are aggregated into a detailed infringement report. This design enhances the transparency and interpretability of the reasoning process;
(2) \textbf{Neural Network Models for Molecular Analysis:} We developed two specialized models -- MarkushMatcher and MarkushParser -- to help LLMs accurately extract substituent groups and recognize Markush structures;
(3) \textbf{MolPatent-240 Benchmark:} To facilitate systematic evaluation, we hire domain experts to curate MolPatent-240, a molecular patent infringement benchmark comprising 240 molecule-patent pairs with diverse and complex Markush structures. 
This dataset reflects real-world patent complexities, constructing a robust foundation for assessing the efficacy of patent infringement algorithms.

Finally, extensive experiments in this paper demonstrate that PatentFinder has achieved significant breakthroughs in the accuracy and interpretability of patent infringement assessments:
(1) PatentFinder achieves state-of-the-art performance, significantly outperforming baseline approaches that rely solely on LLMs. Specifically, PatentFinder surpasses baselines by 13.8\% in F1-score and 12\% in accuracy, demonstrating its effectiveness in handling the nuanced demands of patent infringement assessment;
(2) As illustrated in Figure\ref{fig:main_case_stduy}e, PatentFinder can autonomously generate comprehensive patent infringement reports. 
These reports integrate detailed molecular structure analyses and step-by-step reasoning, resulting in highly informative and easily interpretable outputs. 
The improvements in accuracy and interpretability represent a significant step forward in enhancing the usability of automated patent analysis solutions.
Moreover, our approach of decomposing complex tasks into modular components, each managed by a specialized tool agent, holds potential for application in other workflows of scientific discovery.
\section{Results and Discussion}

PatentFinder introduces two critical designs to facilitate the assessment of small-molecule patent infringement: (1) a multi-agent framework based on LLMs, where various agents equipped with tailored tools address 
manageable subtasks, and (2) two neural network tool models fine-tuned for specific tasks within the patent analysis process, addressing the performance deficiencies of LLMs in these tasks. 
We conduct extensive experiments to demonstrate the effectiveness of these designs.
For the assessment of PatentFinder's multi-agent framework, we compare PatentFinder with state-of-the-art LLMs on small-molecule patent protection discrimination. 
For the fine-tuned tool models, we evaluate them on the tasks of Markush molecular structure matching and Markush image recognition. 
The experimental results demonstrate that these tool models effectively compensate for the deficiencies of general-purpose LLMs in Markush understanding tasks.

\subsection{Evaluation of PatentFinder on MolPatent-240}
\label{sec:main_exp}

\newcolumntype{B}{>{\centering\arraybackslash}p{5em}}
\newcolumntype{C}{>{\centering\arraybackslash}p{6em}}
\newcolumntype{D}{>{\centering\arraybackslash}p{6.5em}}
\begin{table}[ht]
\caption{Patent infringement assessment results on the MolPatent-240.}
\centering
\resizebox{\linewidth}{!}{
\begin{tabular}{l|CCC|DBB}
    \toprule
    \textbf{Method}                & \textbf{F1-Score ($\uparrow$)} & \textbf{Accuracy ($\uparrow$)} & \textbf{Precision ($\uparrow$)} & \textbf{$|$TPR-TNR$|$ ($\downarrow$)} & \textbf{TPR($\uparrow$)} & \textbf{TNR($\uparrow$)} \\
    \midrule
    Random                         & 50.0\%          & 50.0\%          & 43.3\%          & -              & 50.0\%          & 50.0\%          \\
    RDKit Algorithm                & 56.3\%          & 59.0\%          & 49.7\%          & 41.6\%         & 79.8\%          & 38.2\%          \\
    Tanimoto Similarity            & 57.1\%          & 61.0\%          & 50.3\%          & 58.8\%         & 90.4\%          & 31.6\%          \\
    \midrule
    \textbf{\# Patent PDF}\\
    Gemini-1.5-Pro                 & 61.3\%          & 56.1\%          & 72.0\%          & 77.5\%         & 17.3\%          & \textbf{94.9\%} \\
    \midrule
    \textbf{\# Patent  Text + Markush Image}\\
    GPT-4o                         & 57.0\%          & 61.5\%          & 50.3\%          & 69.3\%         & 96.1\%          & 26.9\%          \\
    Claude-3.5-Sonnet              & 61.3\%          & 65.7\%          & 52.8\%          & 66.7\%         & \textbf{99.0\%} & 32.4\%          \\
    Gemini-1.5-Pro                 & 65.8\%          & 68.3\%          & 57.0\%          & 36.5\%         & 86.5\%          & 50.0\%          \\
    \midrule
    \textbf{\# Patent  Text + Markush String}\\
    GPT-4o                         & 64.2\%          & 65.2\%          & 56.7\%          & 15.7\%         & 73.1\%          & 57.4\%          \\
    Claude-3.5-Sonnet              & 65.0\%          & 68.6\%          & 55.6\%          & 53.3\%         & 95.2\%          & 41.9\%          \\
    Gemini-1.5-Pro                 & 70.8\%          & 72.1\%          & 62.5\%          & 19.2\%         & 81.7\%          & 62.5\%          \\
    OpenAI-o1                      & 69.9\%          & 71.2\%          & 61.5\%          & 18.8\%         & 80.6\%          & 61.8\%          \\
    \textbf{PatentFinder}          & \textbf{84.2\%} & \textbf{84.1\%} & \textbf{80.6\%} & \textbf{0.9\%} & 83.7\%          & 84.6\%          \\
    \bottomrule
\end{tabular}
}
\label{tab:main_result}
\end{table}

\subsubsection{Experiment Settings}

To demonstrate the effectiveness of PatentFinder for patent infringement assessment, we construct a patent protection benchmark -- MolPatent-240.
To illustrate the rationale behind our agent-based design, we compare PatentFinder with the direct prompting of leading LLMs, which are tasked with reasoning and determining whether a given molecule infringes on a patent without any additional guidance.
The baselines include state-of-the-art large language models Gemini-1.5-Pro~\cite{gemini} (abbreviated as Gemini), Claude-3.5-Sonnet \footnote{\url{https://www-cdn.anthropic.com/de8ba9b01c9ab7cbabf5c33b80b7bbc618857627/Model_Card_Claude_3.pdf}.} (abbreviated as Claude), GPT-4o~\cite{gpt4o}, and OpenAI-o1 \footnote{\url{https://openai.com/index/introducing-openai-o1-preview}}. 
We leverage the Chain-of-Thought prompts~\cite{CoT} for all the baseline models to enhance their reasoning ability.
As shown in Table~\ref{tab:main_result}, we explore three distinct schemes to assess these models for patent infringement assessment:
(1) \textit{Patent PDF:} 
In this scheme, the patent PDF file is directly provided to the LLM. 
This approach requires the model to have a built-in PDF parsing capability. 
(2) \textit{Patent Text + Markush Image:} 
The model is fed both the textual content of the patent and the core Markush structure images claimed in the patent. 
This scheme requires the model to handle both textual and visual information.
(3) \textit{Patent Text + Markush String:} 
In this case, all Markush images in the patent are first parsed into extended SMILES strings using our MarkushParser model. 
These Markush strings, along with the patent's textual content, are then provided to LLMs in text format, allowing us to utilize text-only models in this scheme.
Since the \textit{Patent PDF} and \textit{Patent Text + Markush Image} schemes require LLMs to handle PDF parsing and image processing, only Gemini, Claude, and GPT-4o are used for these experiments.

We also present the results of several heuristic methods. A structure matching algorithm based on RDKit performs skeleton matching between molecules and Markush structures in patents. If a molecule matches the skeleton of any Markush structure, it is considered to be protected by the patent. Based on the Tanimoto Similarity between the molecular structure and the patent's Markush skeletons, molecules with a similarity greater than 0.5 are regarded as protected by the patent.

In our experiment, we utilize the following evaluation metrics: Micro F1-score, Balanced Accuracy, Precision, True Positive Rate (TPR), and True Negative Rate (TNR). Additionally, We calculate the absolute difference between TPR and TNR for each model to measure its predictive bias towards the True and False classes.
We chose the F1 score as the primary metric for comparison. 
This metric balances precision and recall, ensuring equal emphasis on avoiding false positives and identifying true positives. 

\subsubsection{Results \& Observations}
Experimental results in Table~\ref{tab:main_result} demonstrate that PatentFinder significantly outperforms the direct patent infringement classification with LLMs, regardless of the input scheme adopted.
Specifically, our multi-agent collaboration framework achieves a 12.5\% improvement in F1-score and a 12\% increase in accuracy compared to the best-performing baseline model (Gemini with patent text + Markush string).

These results highlight that instructing LLMs to perform patent assessment without stepwise guidance leads to severe hallucinations and suboptimal performance, which undermines both the reliability and interpretability of LLMs.
In contrast, PatentFinder defines each tool-equipped agent clearly and decomposes the overall complex task into manageable subtasks.
It then organizes multiple agents to address each subtask using the appropriate tools, resulting in a significant performance gain over baselines that use direct prompting methods.

We also observe that using Markush strings generated by MarkushParser instead of the raw Markush images within patents increases the accuracy of the baseline LLMs.
As shown in the table, the patent text + Markush string approaches with Gemini, Claude, and GPT-4o show higher F1 scores (70.8\%, 65.0\%, and 64.2\% respectively) compared to their patent text + Markush image versions (65.8\%, 61.3\%, and 57.0\%).
A similar result is observed when using the raw PDF input with Gemini (70.8\% vs. 61.3\%).
These results suggest that VLMs struggle to interpret Markush structures through the vision modality.
In contrast, our MarkushParser accurately converts Markush images into extended SMILES~\cite{molparser} representations using Optical Chemical Structure Recognition (OCSR), which is more digestible for LLMs whose training data consists predominantly of textual content.

Interestingly, we find that some popular LLMs (\eg GPT-4o and Claude-3.5-Sonnet) tend to adopt an overly conservative classification approach, frequently assuming that the query molecule is protected by the given patent.
They exhibit a TPR that is significantly higher than their TNR, often by 50\% or more. 
Notably, GPT-4o (patent text + Markush string) displays a TPR that exceeds its TNR by 69.3\%. 
Given that MolPatent-240 has a roughly balanced positive-to-negative example ratio (1:1.3), this trend suggests that many leading LLMs tend to classify samples as positive indiscriminately, with some models performing worse than random guessing when distinguishing negative samples (TNR).
In contrast, PatentFinder demonstrates robust performance in distinguishing both positive and negative samples. It achieves a much lower class bias ($|$TPR-TNR$|$=0.9\%), showcasing balanced and accurate classification across the MolPatent-240 dataset.

\subsubsection{Case Studies}
\begin{figure}[t]
    \centering
    \includegraphics[width=\linewidth]{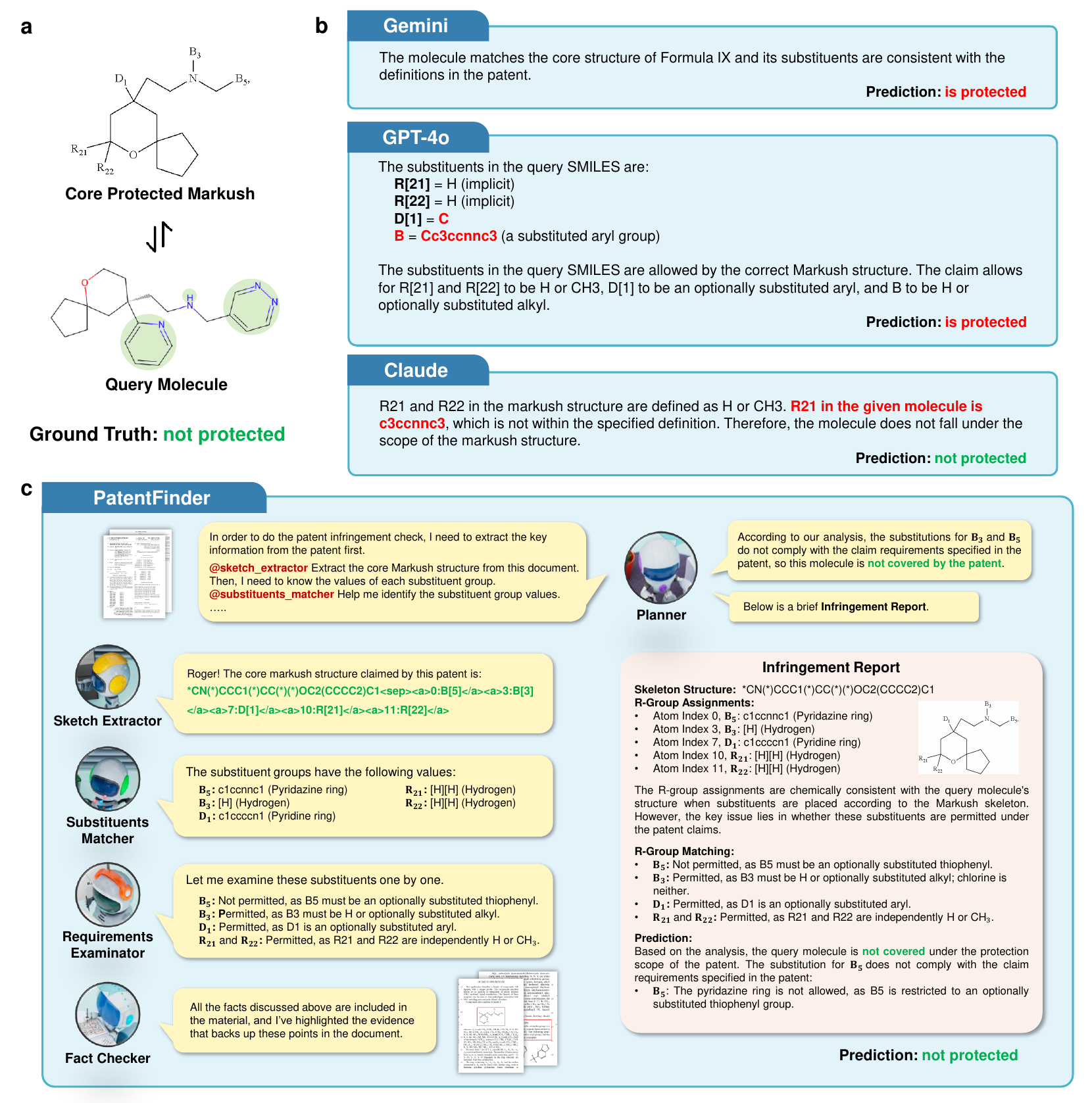}
    \caption{Case study on MolPatent-240. a) A random sample selected from MolPatent-240, the molecule is not protected by the patent. b, c, d) The prediction output of three baseline language models, both implemented with \textit{Patent Text + Markush String} paradigm. e) The inference process of PatentFinder, in which different agents solve the subtasks separately, and their results are summarized into an infringement report. For clarity, some of the outputs are redacted.}
    \label{fig:main_case_stduy}
\end{figure}
In real-world molecular patent infringement assessment, an effective and trustworthy collaboration between models and human experts requires the model to produce human-readable infringement analysis reports. 
These reports should not only deliver a definitive judgment on whether infringement has occurred but also provide a comprehensive and detailed explanation of intermediate reasoning steps. 
To this end, we evaluate not only the accuracy of the infringement assessment but also the validity of reasoning steps across different methods. 
Specifically, we compare the intermediate outputs of baseline models and PatentFinder on a real example from the MolPatent-240 dataset. 
The results are illustrated in Figure\ref{fig:main_case_stduy}. 
For baseline models, we summarize the step-by-step reasoning and final decisions of Gemini, GPT-4o, and Claude, while for PatentFinder, we present the intermediate outputs from each agent and the final infringement report generated by the Planner based on these analyses.

As shown in Figure\ref{fig:main_case_stduy}b, directly using LLMs for patent protection determination often leads to hallucinations in intermediate reasoning steps, resulting in incorrect conclusions based on non-existent evidence. 
For instance, Gemini provides no intermediate analysis and arrives at an incorrect judgment after merely comparing the molecular skeletons to the Markush structure. 
GPT-4o attempts to match the skeleton against the Markush structure but misidentifies the substituent D1 and introduces a fictitious substituent B. 
Although Claude reaches the correct conclusion, its analysis erroneously identifies R[21] as a chlorine atom and fails to determine the values of three other substituents.

From the baseline results, two key observations emerge: 
(1) Only GPT-4o attempts a fine-grained comparison of all substituents, while the other two models deliver conclusions without detailed analysis.
(2) LLMs exhibit limited capabilities in comparing chemical structures, with reasoning errors propagating through subsequent analysis steps.

PatentFinder addresses these challenges by employing a multi-agent framework that decomposes the complex task of patent infringement detection into a series of simpler subtasks. 
Each subtask is more comprehensible to a tool agent, enabling fine-grained analysis at every step. 
Additionally, the integration of domain-specific tools compensates for the shortcomings of pure LLM approaches in chemical structure analysis. 
This design significantly reduces the occurrence of hallucinations and reasoning errors. 
At the final stage, the Planner aggregates the outputs from all agents to generate a detailed infringement analysis report that includes a robust reasoning process.

\subsection{Evaluation of MarkushMatcher}
\begin{figure}[t]
    \centering
    \includegraphics[width=\linewidth]{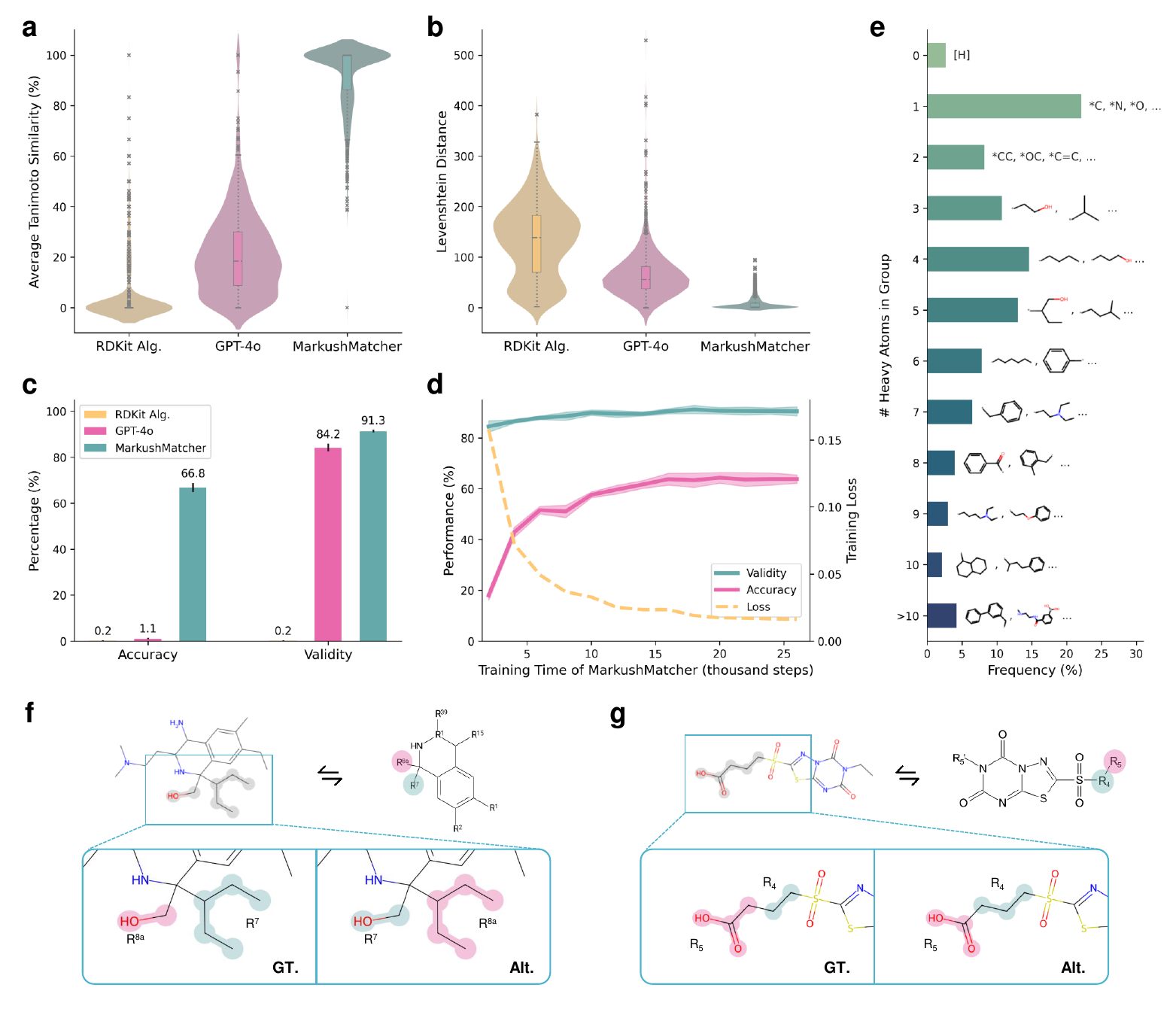}
    \caption{\textbf{Performance comparison between different substituent group extraction methods.} a) Average Tanimoto similarity of different matching algorithms. b) Levenshtein distances of different matching algorithms. c) Exact match accuracy and chemical validity of different matching algorithms. d) The accuracy, validity, and training loss of MarkushMatcher at different training steps. e) Statistics of the substituent groups used in training MarkushMatcher, where the groups with a single heavy atom (C, N, O) appear most frequently. f, g) The exchangeable groups and adjacent groups make the substituent group matching task unsolvable without additional textual description, which indicates the necessity of including contexts in the patent. GT: Ground Truth; Alt: Alternative solution.}
    \label{fig:markush_matcher_exp}
\end{figure}
\begin{figure}[t]
    \centering
    \includegraphics[width=\linewidth]{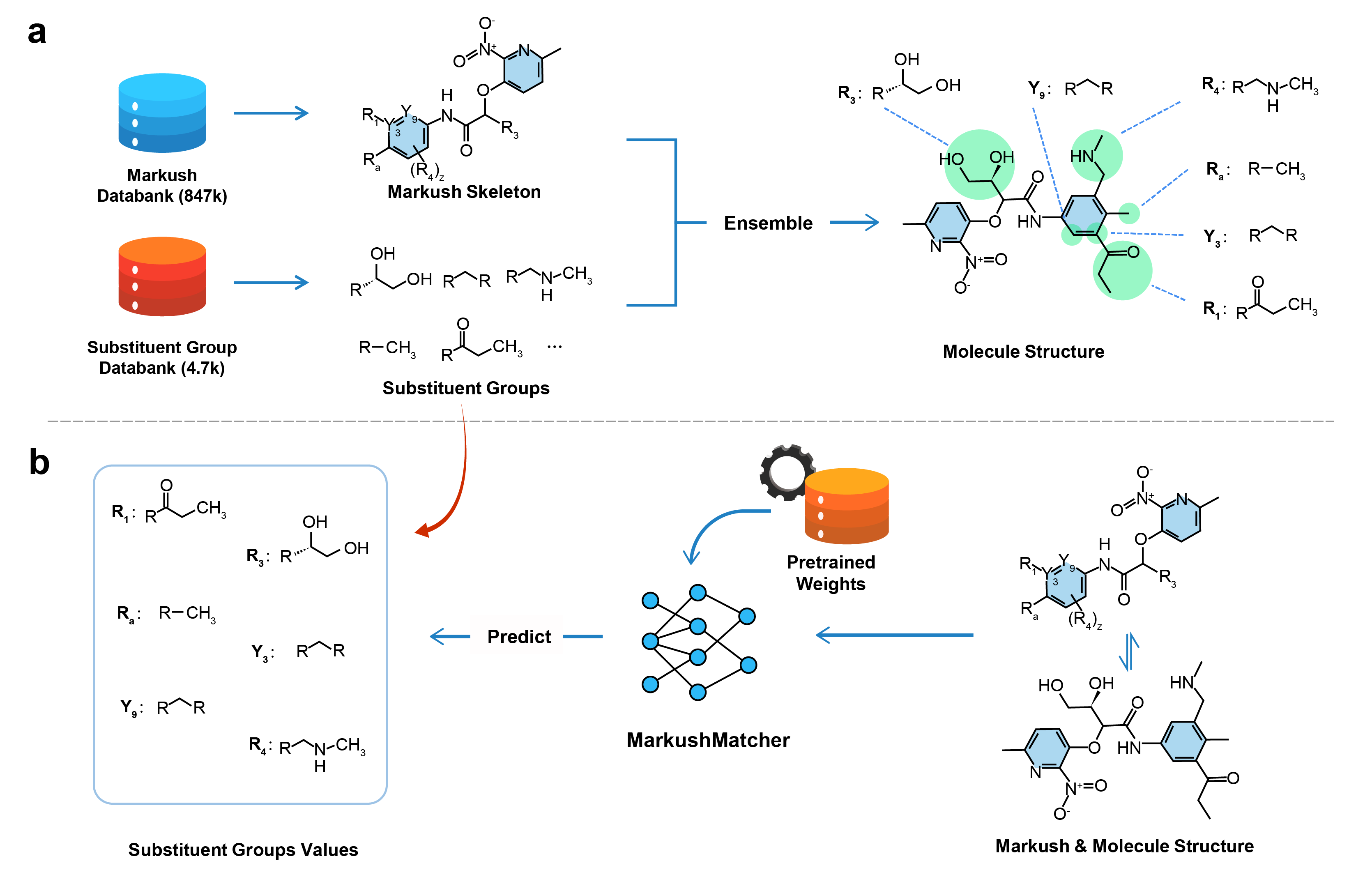}
    \caption{\textbf{Illustration of the development of MarkushMatcher}. a) We construct Markush matching data by designing a reverse data generation algorithm by attaching the substituent groups onto the Markush Skeletons. b) During training, the MarkushMatcher model takes the Markush structure and the query molecule as input, and is trained to predict the values of each substituent group defined in the Markush structure.}
    \label{fig:markush_matcher}
\end{figure}

To showcase the ability of MarkushMatcher to deliver precise structural matching results, we curate a dataset comprising 1,000 samples for substituent extraction. 
The comparison focuses on four key metrics: 
(1) \textbf{Accuracy:} The proportion of predictions that perfectly match the true substituent results. 
Perfect matches are defined as having a Tanimoto~\cite{tanimoto_best_choice, tanimoto} similarity of 100\%.
(2) \textbf{Validity:} The proportion of valid predictions. 
A prediction is considered valid if it includes all substituents mentioned in the Markush structure and each predicted substituent corresponds to a legitimate molecular structure.
(3) \textbf{Average Tanimoto Similarity:} The average Tanimoto Similarity between each predicted substituent and its corresponding ground truth substituent, computed using Morgan Fingerprints~\cite{morgan-fingerprints} for all substituents in a sample, the higher the better.
(4) \textbf{Levenshtein Distance:} The distance between the predicted results and the ground truth is calculated directly at the string level, the lower the better.

We compare MarkushMatcher with structure-matching baselines implemented using the rule-based -- RDKit~\cite{rdkit} and LLM-based -- GPT-4o methods, respectively.
As shown in Figure~\ref{fig:markush_matcher_exp}a-c, MarkushMatcher outperforms both baseline methods on the substituent groups prediction task. 
As a deterministic algorithm, the rule-based approach -- RDKit struggles with complex substituent matching tasks. 
While effective for simple tasks involving single-link substituents, it performs poorly in more intricate scenarios. 
This limitation result in near-zero accuracy and validity scores on our test set (\cf Figure~\ref{fig:markush_matcher_exp}a,b), with unsatisfactory results for the other two metrics as well.
Using LLMs alone does not yield satisfactory results either. 
Although GPT-4o is able to provide chemically valid solutions in many scenarios, which results in the high validity in Figure~\ref{fig:markush_matcher_exp}c, it fails to predict the correct substituent group values in most cases, leading to low similarity, high Leveshtein distance, and limited accuracy. Common errors include: a) Confusing the values of different substituents; b) Incorrectly identifying substituent positions on the Markush structure; and c) Misclassifying scaffold atoms as part of the substituents.
MarkushMatcher outperforms the other two methods by a large margin, with an accuracy of 66.8\%. 
In comparison with GPT-4o, MarkushMatcher increases the mean average Tanimoto similarity from 21.3\% to 92.9\% and has an 8.8 times lower average Levenshtein distance.
The training details of MarkushMatcher are presented in Figure~\ref{fig:markush_matcher_exp}d, e. 
During the model's training, the training loss converges eventually, with a steady increase in both accuracy and validity.
In the training data of MarkushMatcher, monatomic substituents such as C, N, and O are the most frequently occurring, while the heavy atom counts of other groups follow an approximately bell-shaped distribution.

We also find that in certain cases, even with the given Markush and molecular structures, it is impossible to determine a unique result without additional contextual information. 
For instance, the presence of exchangeable groups (\cf Figure\ref{fig:markush_matcher_exp}f) and adjacent groups (\cf Figure\ref{fig:markush_matcher_exp}g) in the Markush structures can introduce complexities. 
Specifically, if two substituent groups are connected to the same atom, the correlation between the group identifiers and the substructures may become ambiguous. 
Similarly, in cases where multiple substituent groups are adjacent, there could be various valid allocation schemes when specifying the group values.

In case these situations are commonly presented in the real-world patent Markush structures, we retain these data during the evaluation of substituent group extraction methodologies. 
Under these conditions, MarkushMatcher achieves a 92.9\% Tanimoto Similarity, a remarkably high result considering the dataset’s complexity.
Consequently, we employ the MarkushMatcher solely as a tool model within the PatentFinder pipeline, ensuring that the outputs of this model are meticulously validated by agents based on the complete content of the patent.

\subsection{Evaluation of MarkushParser}

\begin{figure}
    \centering
    \includegraphics[width=.95\linewidth]{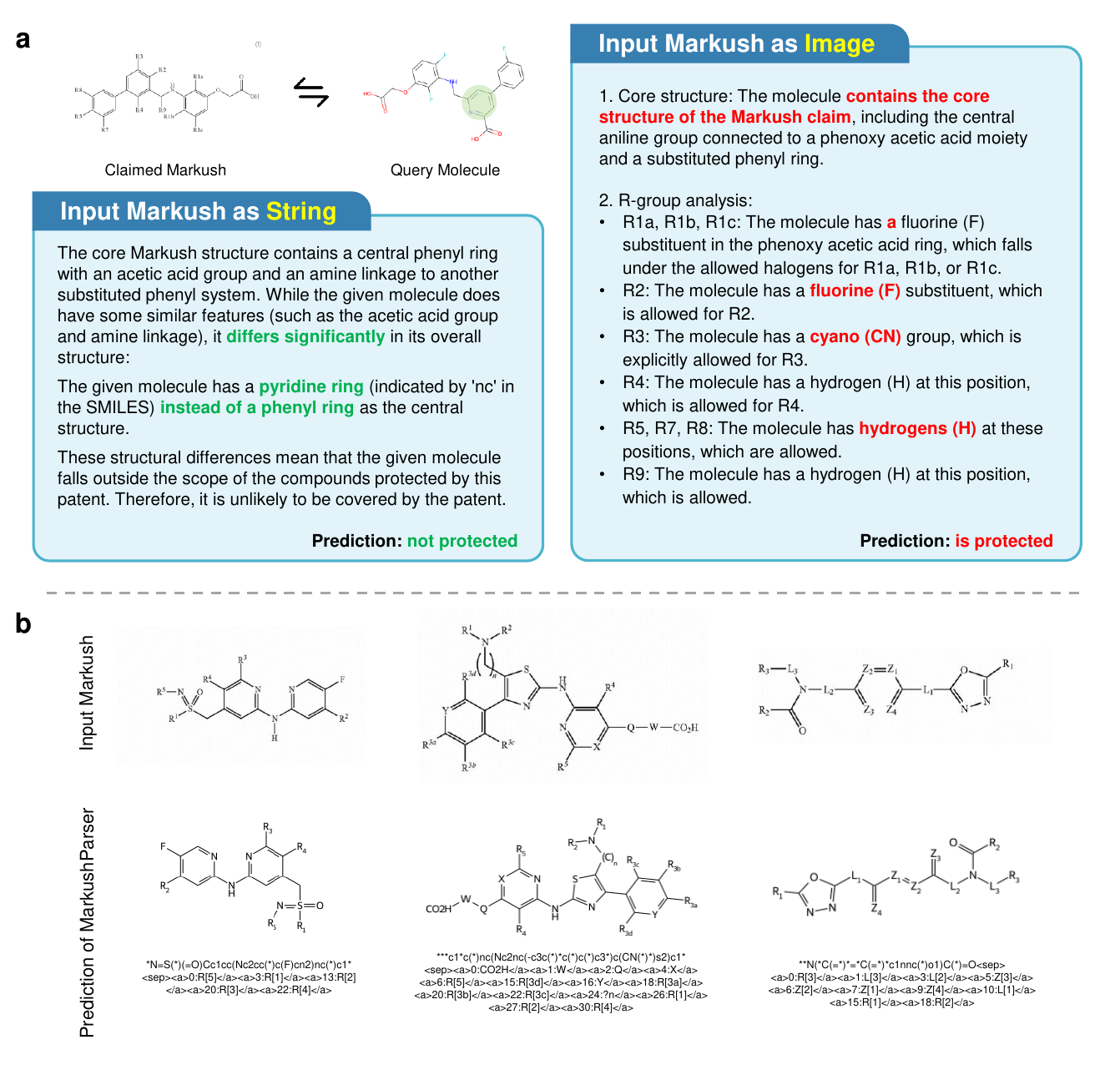}
    \caption{\textbf{Case Studies of MarkushParser.} a) Comparison between different Markush input modes: Markush string and Markush image. Both groups use Gemini as the Backbone Language model. b) Markush structure reconstruction results of MarkushParser.}
    \label{fig:markush_parser_exp}
\end{figure}

To demonstrate the effectiveness of MarkushParser in parsing Markush images into strings, we conduct experiments on MolPatent-240. 
We maintain the condition that the query molecules are always input in SMILES format and compare three different Markush input methods: original patent PDFs, Markush images, and Markush strings parsed by MarkushParser. 
We utilize three LLMs that support vision inputs -- GPT-4o, Claude, and Gemini -- to meticulously compare their intermediate outputs and the process of chain-of-thought reasoning under the different input modalities. 
The experimental results are shown in Table~\ref{tab:main_result} and Figure\ref{fig:markush_parser_exp}a. 
The figure illustrates that the LLMs struggle to accurately understand the structure of Markush under the image and PDF understanding schemes, often erroneously fabricating Markush structures and non-existent substituent groups for analysis.


In Figure\ref{fig:markush_parser_exp}b, we present several results of the MarkushParser's parsing of Markush images. The first row in the figure displays the original input Markush images, while the second row shows the extended SMILES strings generated by the MarkushParser and the reconstructed Markush structures based on these parsing results. The MarkushParser effectively captures the structural details of the Markush structures, and the reconstructed outputs exhibit high consistency with the original inputs.
\section{Conclusion}

PatentFinder drives automated drug discovery forward by tackling the critical challenge of patent infringement assessment for small molecular drugs. 
Through its novel tool-enhanced and multi-agent framework, PatentFinder addresses the inherent limitations of LLMs in interpreting complex molecular structures and Markush notations with precision and reliability.

Rigorous evaluations on the MolPatent-240 dataset highlight PatentFinder’s superior performance, achieving enhanced accuracy, balanced predictions, and improved interpretability. 
By autonomously generating detailed infringement reports, PatentFinder ensures transparency and usability in its assessments.
It fills a crucial gap in the automated drug discovery pipeline and enables seamless integration of patent compliance checks into pharmaceutical workflows, thus paving the way for more efficient and compliant pharmaceutical innovation.
\section{Methods}

\subsection{Data Construction}
This section outlines the data processing procedures applied to all datasets used in this paper.
\subsubsection{Construction of MolPatent-240}
\label{sec:molpatent-240}
Due to the lack of publicly available molecular patent infringement datasets, we manually construct MolPatent-240 to facilitate evaluation. 
This dataset comprises 240 patent–molecule pairs, each labeled as either ``protected'' or ``not protected'' by human experts.

To construct this dataset, we collect 70 molecular protection patents from public online sources.
For positive examples, we select existing embodiments from the patents or manually designed novel molecules based on the patent claims. 
For negative examples, we modify the structures of positive examples at the atomic level, introducing subtle differences in the molecular skeleton or substituents that deviate from the patent claims.

After construction, three human experts specializing in drug discovery verify the chemical correctness of all samples. 
This includes validating the legality of the constructed molecules and the correctness of their protection labels. 
Based on their feedback, individual dataset contributors revise the benchmark to correct errors and remove a subset of samples that are too challenging to rectify.

\subsubsection{Construction of Substituent Groups Extraction Data}
\label{sec:data_markush_Matcher}
Training MarkushMatcher requires large-scale data, which is impractical to annotate manually with human experts. 
Furthermore, the task of matching Markush structures with molecules and extracting substituent groups is exceptionally challenging, as rule-based algorithms and LLMs fail to produce satisfactory results. 
To address this, we propose a reverse data construction method that samples Markush structures and substituent groups from large-scale databases, assembles them into complete molecules using chemical tools, and records the substituent matching relationships during the assembly process. 
The entire pipeline is depicted in Figure\ref{fig:markush_matcher}.

MolParser-7M~\cite{molparser}, a dataset containing over 7 million molecular and structural images, serves as the foundation for this process. 
Some samples in MolParser-7M include Markush structures. 
We heuristically design filtering rules to exclude unsuitable samples, such as those without Markush structures, those containing multiple molecules, or molecules that are excessively small. 
This filtering process results in a dataset of 847 thousand Markush structure records.
We manually curate and organize a substituent group database containing 4.7 thousand substituents derived from molecular structures.
The distribution of the substituent groups in the training set is shown in Figure\ref{fig:markush_matcher_exp}e.
For each Markush sample in the dataset, we randomly sample substituent groups from this database according to the number of substituent attachment points on the scaffold. 
Using RDKit, we combine the Markush scaffold with the sampled substituents based on predefined rules to generate complete molecules. 
The substituent matching relationships are used as supervisory signals during the model's training procedure.

Through this reverse construction method, we generate a dataset containing 847 thousand pieces of structural matching data. 
We randomly sample 1,000 examples from this dataset for model evaluation, with the results shown in Figure~\ref{fig:markush_matcher_exp}. 
The remaining samples are used to train MarkushMatcher.

\subsubsection{Construction of Markush OCSR Data}
\label{sec:data_markush_ocsr}
To facilitate the parsing of Markush diagrams in the patent, we collect 515 real-world Markush images and manually label their corresponding string expressions.
During labeling, we employ the extended SMILES notation defined in~\cite{molparser} to represent Markush structures as text.
This dataset includes hand-drawn molecular images, Markush structures extracted from molecular protection patents, and other noisy samples. 

The annotation process is conducted collaboratively by several researchers with expertise in medicinal chemistry, who meticulously label the Markush strings corresponding to each molecular image. 
This dataset serves as a valuable resource for finetuning MolParser to handle complex and noisy Markush recognition tasks.

\subsection{LLM-based Agents}
We have designed multiple LLM-based agents to tackle subtasks, each fulfilling distinct roles:

\textbf{Planner.}
As illustrated in Figure~\ref{fig:framework}, this agent interacts directly with users by receiving input queries (i.e., patent and molecule) and providing output results (i.e., infringement report). 
The Planner collects the necessary information and determines which agent to engage next. 
Each time it decides to invoke an agent, it generates an instruction in the form of a triplet: (thought, target agent, input to agent).
After sending the instruction to the target agent, the Planner awaits the ``agent response'' and continues to formulate new triplets based on the ongoing chat history.
Once it determines that sufficient information has been gathered, it presents its infringement judgment to the user.

\textbf{Sketch Extractor.}
This agent is responsible for extracting the protected molecular skeletons from the patent. 
It supports reading patent content from both online sources and local PDF files with Patent Web Search and deployed parsing tool, respectively.
It identifies key protected Markush structures and the associated claim requirements, specifically the substituent group values outlined in the patent.

\textbf{Substituents Matcher.}
This agent assesses the structural similarity between the query molecule and the extracted molecular skeleton to determine skeleton matches.
It is equipped with a substructure matching algorithm and a substituent group extraction model. 
Both results from the matching algorithm and the extraction model are provided as references to this agent.
Using vision-LLM (GPT-4o with image input), this agent verifies substituent group mappings, corrects chemical errors made during the substituent group extraction, and performs necessary rearrangements to the mappings based on the requirements.

\textbf{Requirements Examinator.}
This agent compares the query molecule with the requirements defined in the patent, conducting the analysis at both the skeleton level and the substituent group level.
At the skeleton level, the agent assesses the query molecule against the Markush structural skeleton and relevant requirements outlined in the patent (e.g., the overall number of heavy atoms or whether the molecule is explicitly discussed in the patent).
For the substituent group level comparison, the agent evaluates the values of the substituent groups on the molecule against the definitions provided in the patent.
After completing these comparisons, the agent reports the results to the Planner.

\textbf{Fact Checker.}
This agent ensures that each requirement argument used in the previous reasoning steps is indeed present in the patent document.
If the molecule is determined to be protected by the patent, the agent further identifies the core Markush structure that safeguards the molecule and locates the associated textual claim requirements. 
Subsequently, the agent utilizes the document segmentation model to divide the patent document and highlight the relevant claim requirement elements in the original PDF file for the user's reference.

In our framework, we utilize two distinct LLMs: OpenAI-o1, which functions as the Requirement Examiner with a temperature setting of 1.0 to encourage reasoning, and GPT-4o, employed for other agents with a more conservative temperature setting of 0.2 to ensure consistent and accurate outputs.

\subsubsection{Neural Network Tool Models}
\label{sec:tool-model}

\textbf{MarkushMatcher.}
The MarkushMatcher is a neural network model designed to predict the specific values of various substituents as defined by a given Markush structure in a queried molecule, based on the Markush structure and the molecular query. 

MarkushMatcher is built upon the T5~\cite{t5}, an encoder-decoder transformer~\cite{transformer} architecture, and is initialized with a MolT5-large~\cite{molt5} checkpoint.
The training process of MarkushMatcher is illustrated in Figure~\ref{fig:markush_matcher}b. 
To adapt the model for substituent group prediction, we fine-tune it using a large corpus of generated data. 
The input comprises molecular and Markush structures expressed as string representations, while the output, which maps substituent group values, is represented in JSON format.
As detailed in Figure~\ref{fig:markush_matcher}a, we construct a dataset of 847 thousand substituent group prediction samples for training and evaluation. The fine-tuning process involves training the model for one epoch on this dataset, using a batch size of 8 and a learning rate of $1 \times 10^{-4}$.

\textbf{MarkushParser.}
MarkushParser converts molecular images into extended SMILES representations with OCSR (Optical Chemical Structure Recognition) and can translate common molecules, substituent groups (incomplete molecules with connection points), and Markush structures. 

This model uses Swin Transformer~\cite{swin_transformer} as the image encoder and applies a two-layer multiple-layer perceptron as a vision-language connector similar to LLaVA~\cite{llava}. After the vision encoder and connector, a BART model~\cite{bart} decodes the compressed image features to predict the extended SMILES as text sequences.

We manually collect and label 515 high-quality Markush images and their associated extended SMILES strings from real-world patent or drug discovery scenarios. The model is first initialized with MolParser~\cite{molparser}, then fine-tuned with our 515 Markush image-string pairs for 20 epochs with a learning rate of $5\times 10^{-5}$.

\textbf{Document Segmentation.}
The document segmentation tool, which is instantiated with a YOLO object detection model, recognizes all the elements in the PDF pages and locates their coordinates on the page. 
Each element could be a Markush/Molecular image, a table, or a paragraph in the PDF. 

\subsubsection{Heuristic Tools}

\textbf{Patent Web Search.}
This function retrieves segmented molecular images and plain text content from the Google Patents database~\footnote{\url{https://patents.google.com/}}.
It is particularly useful for handling input queries when only the patent ID is provided.
Notably, our framework remains fully functional without internet-based support, as it can alternatively rely on local PDF parsing and image processing through the OCSR scheme. 
This flexibility ensures that users can still perform comprehensive analyses even when web-based resources are unavailable.

\textbf{PDF Parser.}
This tool converts the PDF patent document into images and texts, facilitating segmentation and extraction of key sections.

\textbf{RDKit Substructure Matcher.}
We implement a substructure extraction algorithm with RDKit~\cite{rdkit} to match the query molecule with the extracted Markush structure. 
Given the SMILES representation of a query molecule, this tool extracts the values of each substituent group defined in the Markush.
\section*{Data availability}
During the training data construction of MarkushMatcher, we use the open-source Markush databank at \url{https://huggingface.co/datasets/AI4Industry/MolParser-7M}. Our patent infringement assessment benchmark MolPatent-240 is released at \url{https://github.com/syr-cn/patentfinder_code_private/tree/master/patent_finder/data}.

\section*{Code availability}

The code of PatentFinder is available at \url{https://github.com/syr-cn/patentfinder_code_private}, including the code of the multi-agent inference framework, implementations of the heuristic tools, and neural network tools. The 847 thousand constructed data for training MarkushMatcher and the trained model parameters are available at \url{https://osf.io/ftq9g}. The 515 human-labeled data used for fine-tune MarkushParser are available at \url{https://osf.io/jsqa3}.

\section*{Corresponding authors}
Correspondence authors: Xiang Wang and Hengxing Cai.

\section*{Competing interests}
The authors declare that they have no known competing financial interests or personal relationships that could have appeared to influence the work reported in this paper.

\newpage
\bibliographystyle{nature}

\bibliography{references}  





\newpage

\appendix

\section{Implementation Details}

Within the PatentFinder framework, LLMs are assigned distinct roles and collaborate systematically to address the task of patent infringement analysis. 
During the process, input prompts are carefully tailored to optimize the models’ adaptation to their respective roles. 
This section outlines the role-specific prompts designed to guide the cognitive processes of the models.

Initially, each agent is equipped with the foundational information necessary to comprehend the definition and structure of extended SMILES strings.

\begin{tcolorbox}[title = {Extended SMILES Definition}]
\scriptsize
\begin{lstlisting}
We define a markush string as: `SMILES<sep>EXTENSION`
where:
- `SMILES` is the basic molecular skeleton
- `EXTENSION` is a list of completions for the R-groups in the SMILES string, such as R-group definitions, substitutions, or modifications.
- The EXTENSION is optional and is in XML format.
    - `<a>[ATOM_INDEX]:[GROUP_NAME]</a>` indicates that the atom at index ATOM_INDEX (which is the index of the corresponding asterisk in the SMILES string) in the SMILES string corresponds to the R-group named GROUP_NAME, starting from 0. In this case, the number of asterisks in the SMILES string should be equal to the number of <a> tags in the EXTENSION.
    - <r>[RING_INDEX]:[GROUP_NAME]</r> indicates that the ring at index RING_INDEX, which is the index of the rings in the SMILES string in the order of appearance. For example, the first-occurred ring has an index of 0.
    - <c>[CIRCLE_INDEX]:[CIRCLE_NAME]</c> indicates that the circle at index CIRCLE_INDEX in the SMILES string corresponds to the R-group named CIRCLE_NAME.
    - A special token <dum> is used to indicate a connection point in the SMILES string.
    - GROUP_NAME is the name of the R-group, which can be an abbreviation or full name (e.g., R, X, Y, Z, Ph, Me, OMe, CF3, etc.).
    - If the GROUP_NAME is a markush substituent group and there's a subscription, the subscript is appended to the GROUP_NAME with a colon (e.g., R[1], R[2], R[3], etc.).
    In the pdf parsing results, the OCR results of markush and molecular SMILES images are wrapped in a \\caption tag.
\end{lstlisting}
\end{tcolorbox}

Subsequently, each intelligent agent is equipped with a carefully crafted, human-written prompt tailored to its specific role within the system.

\begin{tcolorbox}[title = {Prompt for Sketch Extractor}]
\scriptsize
\begin{lstlisting}
You are an expert in analyzing chemical patent documents. Your task is to identify the sections of a patent that contain key information about a specific Markush structure and its associated claim requirements.

# Task Overview:
- Read through the patent document provided.
- Identify and mark the beginning and end indices of sections that discuss the core Markush structure and its detailed claim requirements.
- These sections are critical as they determine the scope of patent protection for the molecule.

# Special Instructions:
- The core Markush structure is often the first structure mentioned in the patent.
- Extract the Markush structures, not the molecule examples or embodiments. The Markush structures always have a <sep> tag.
- Pay close attention to descriptions of the molecular structure, especially the parts that detail variations and substitutions allowed under the patent.
- Highlight any discussions related to the legal scope of the patent, including examples, embodiments, and specific conditions mentioned for the Markush groups.
- Try to be concise. No more than {max_blocks} blocks after summarization are tolerable for this task.

Input the full patent text below and use the provided Markush structure as a reference to guide your extraction process.

{EXTENDED_SMILES_DEFINITION}

# Patent PDF Blocks:
{block_text}
\end{lstlisting}
\end{tcolorbox}

\begin{tcolorbox}[title = {Prompt for Substituents Matcher}]
\scriptsize
\begin{lstlisting}
You are an expert in chemical patents and molecular representations. Your task is to verify the correctness of the substructure match result between the markush claim and the query molecule.

{EXTENDED_SMILES_DEFINITION}

The R-group mapping is a dictionary that maps the R-group names to the corresponding SMILES strings. The R-group names are the keys, and the SMILES strings are the values. These mappings are extracted during the match between a markush string and a molecule instance.

Example 1:
- Markush: *c1*c(*)c2cc(*)c(*)cc2n1<sep><a>0:R[1]</a><a>2:X</a><a>4:R[a]</a><a>8:R[2]</a><a>10:R[3]</a>
- Molecular SMILES: c1(C(=O)O)cc(O)nc2ccc(C#N)cc12
- R-Group Mapping: {"R1": "O", "X": "C", "Ra": "C(=O)O", "R2": "C#N", "R3": "[H][H]"}

Example 2:
- Markush: *C(C(=*)N(*)*)Sc1nc2*:*:*:*c2c(=O)n1*<sep><a>0:R[2]</a><a>3:A[5]</a><a>5:R[4]</a><a>6:R[3]</a><a>11:A[4]?n</a><a>12:A[3]</a><a>13:A[2]</a><a>14:A[1]</a><a>19:R[1]</a>
- Molecular SMILES: C(C)(Sc1nc2ccccc2c(=O)n1C1CCCC1)C(=O)N(C)c1ccccc1
- R-Group Mapping: "substructure_map": {"R2": "C", "A5": "O", "R4": "c1ccccc1", "R3": "C", "A4": "C", "A3": "C", "A2": "C", "A1": "C", "R1": "C1CCCC1"}

Example 3:
- Markush: *C(=O)C(*)(*)Cc1ccc(C(=O)Oc2ccc(C(=N)N)cc2*)s1<sep><a>0:X</a><a>4:R[1]</a><a>5:R[2]</a><a>23:R[7]</a>
- Molecular SMILES: CC(C)(Cc1ccc(C(=O)Oc2ccc(C(=N)N)cc2F)s1)C(=O)Nc1cc(C(=O)O)cc(C(=O)O)c1
- R-Group Mapping: {"X": "Nc1cc(C(=O)O)cc(C(=O)O)c1", "R1": "C", "R2": "C", "R7": "F"}

Note that: 
- If you don't think the molecular matches the claim, return `r_group_matching` as an empty dictionary.
- The R-group values are considered correct iff when we replace the R-groups in the Markush structure with the corresponding values, we get the query molecule. Otherwise, the R-group values are incorrect.
- Be very cautious about any potential protection scope of the patent. If this sample is miss classified, it may lead to a serious legal issue.
- The substructure match should only be used as a reference, since the matching algorithm may not be perfect.
- In your reasoning, compare each R-group definition with the query molecule, and analyze the correctness of the R-group values one by one.

After your analysis, provide a verified R-group mapping result explicitly.


**Markush Structure**: `{markush_string}`

**Query Molecule**:
{mole_string}

**R-Group Mapping Extracted by Chemical Software**:
{r_group_mapping}

**R-Group Mapping Extracted by neural network model**:
{nn_result}
\end{lstlisting}
\end{tcolorbox}
\newpage

\begin{tcolorbox}[title = {Prompt for Requirements Examinator}]
\scriptsize
\begin{lstlisting}
You are an expert in chemical patents and molecular representations. You are tasked with determining whether a given molecule is covered under the protection scope of a specific patent.
{EXTENDED_SMILES_DEFINITION}

Special Attention Required:
- In your analysis, pay particular attention to the matching of R-group values from the substructure match result with the stipulations outlined in the claim requirement text. It is crucial to meticulously assess whether each R-group and its corresponding substitutions align with the conditions specified for patent coverage. Ensure that this careful examination is clearly reflected in your step-by-step reasoning.
- Though the molecule's skeleton may match the markush, you must verify the value of each R-group value by comparing them against the claim requirement text one by one. Also, mention the corresponding definition of each R-group in your analysis.
- Be careful to draw a "is protected" conclusion. If any of the R-group values are not protected by the claim, the molecule is not protected by the patent.
- If you do find out every R-group value is protected by the claim, don't be afraid to draw a "is protected" conclusion. But make sure to provide a clear reasoning for your conclusion.

# Output Expectation:
- **Analysis:** Carefully analyze the R-group substitutions in the query molecule based on the claim requirement text. Ensure that each R-group and its corresponding substitutions align with the conditions specified for patent coverage.
- **Prediction:** Provide a definitive conclusion on whether the query molecule is covered under the protection scope of the patent.

**Markush Claim**: `{markush_string}`

**Current Substructure Match Result:**
{structure_match}

**Query Molecule**:
{mole_string}

**Claim Requirement Text:**
{claim_text}
\end{lstlisting}
\end{tcolorbox}

\begin{tcolorbox}[title = {Prompt for Fact Checker}]
\scriptsize
\begin{lstlisting}
You are an expert in chemical patent analysis with a focus on molecular structures and patent claims.

# Task Overview:
- Check the reasoning detail of the previous steps in a patentability analysis process, and make sure that all of the evidence used in these steps has appeared in the raw patent PDF.
- Extract which of the provided blocks contain the markush structure that actually protects the target molecule.
- Extract the key sections of the patent that have directly contributed to the infringement analysis.
- Direct contributions include R-Group definitions infringement, and other relevant claim requirements.
- For each extracted block, explain why is it relevant to the infringement analysis by summarizing the key fact claimed in the block.
- Then length of requirement_indices and requirement_details should be the same. Each element in the list `requirement_details` should correspond to the block index in `requirement_indices`.

# Special Instructions:
- Focus on the detailed claim requirements that specify the allowed variations, substitutions, or conditions for the Markush groups.
- Be as precise as possible when locating the most relevant blocks.
- Try to be concise. You should identify the most relevant {max_blocks} blocks that directly contribute to the infringement analysis.
- Return as few requirement block indices as you can. If one block is enough to provide a clear answer, only one block is needed.
- Sort the block indices by their contribution to the infringement analysis.

{EXTENDED_SMILES_DEFINITION}


# Target Molecule:
{target_smiles}

# Patent PDF Blocks:
{block_text}

# Infringement Analysis Results:
Infringement: {input_is_protected}
Analysis: {input_reasoning}
\end{lstlisting}
\end{tcolorbox}
\newpage

\begin{tcolorbox}[title = {Prompt for Planner}]
\scriptsize
\begin{lstlisting}
You are an expert in chemical patents and molecular representations. You are tasked with determining whether a given molecule is covered under the protection scope of a specific patent.
{EXTENDED_SMILES_DEFINITION}

%Prompt for Planning%
You will be provided with some information about a molecule and a relevant markush patent. Your task is to make plans for other agents, 
- Skeleton Extractor extracts the core Markush structures and associated claim requirements from the Patent. Parameters: (blocks: list[dict])
- Substituents Matcher identifies and validates substituent groups in Markush expressions relative to the query molecule. Parameters: (markush_string: str, molecule_string: str, rdkit_result: dict, nn_result: dict, patent_text: str)
- Requirements Examinator assesses whether the query molecule meets the patent's substituent group requirements. Parameters: (markush_string: str, molecule_string: str, match_result: dict, patent_text: str)
- Fact Checker verifies agents' outputs against original claims and corrects discrepancies for accuracy. Parameters: (target_smiles: str, blocks: list[dict], input_is_protected: bool, input_reasoning: str)

%Prompt for Summarizing%
Some neural network models have already been used to match the molecule with the markush claim and analyze the R-group substitutions in the query molecule based on the claim requirement text.
Your task is to carefully re-organize the information into a comprehensive infringement report, and provide a definitive conclusion on whether the query molecule is covered under the protection scope of the patent.

Your infringement report must include:
- The structure of the query molecule.
- The structure of the core Markush protected in the patent, and the corresponding R-group definitions.
- The value of each R-group is defined in the Markush claim.
- Whether the substitutions of each R-group in the query molecule align with the conditions specified for patent coverage.
- Whether there are any other requirements in the patent, which the molecule infringes.
- A definitive conclusion on whether the query molecule is covered under the protection scope of the patent.
\end{lstlisting}
\end{tcolorbox}

\newpage
\section{Statistics of MolPatent-240}

MolPatent-240 consists of 240 molecule-patent pairs, including 104 positive examples (protected samples) and 136 negative examples (non-protected samples). Positive samples are either directly selected from patent embodiments or manually constructed to meet specific patent requirements. Negative samples are created by violating patent substituent requirements, altering the core Markush skeleton, or selecting embodiments from unrelated patents. Table~\ref{tab:ds_statistics_1} summarizes the number of samples for each label type and the corresponding construction approach.

\begin{table}[h]
\caption{Overview of dataset construction methods categorized by label type.}
\centering
\resizebox{0.9\linewidth}{!}{
\begin{tabular}{CCl}
\toprule
Label      & \#Samples & Construction Method                                                      \\
\midrule
Positive   & 67        & Selected from embodiments in the patents.                                \\
Positive   & 37        & Manually constructed based on the patent requirements.                 \\
Negative   & 73        & Manually constructed by violating the patent substituent requirements. \\
Negative   & 20        & Manually constructed by violating the core Markush skeleton.             \\
Negative   & 43        & Selected from embodiments in unrelated patents.     \\
\midrule
Total      & 240       & - \\
\bottomrule
\end{tabular}
}
\label{tab:ds_statistics_1}
\end{table}

\begin{figure}[h]
    \centering
    \includegraphics[width=0.7\linewidth]{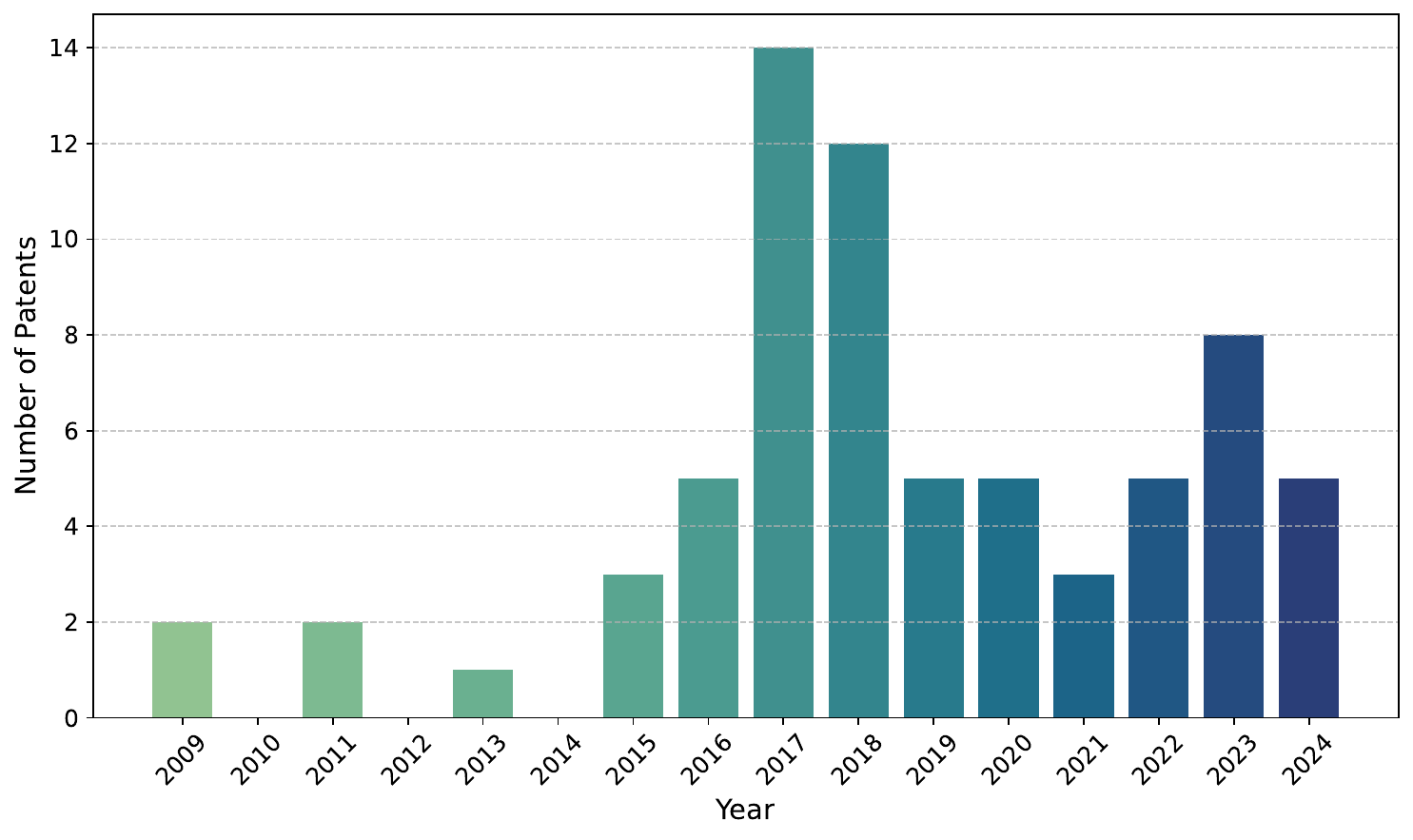}
    \caption{The publication years of patents in MolPatent-240.}
    \label{fig:statistics_1}
\end{figure}

The dataset comprises 70 unique patents, with 9 sourced from WIPO (World Intellectual Property Organization, prefixed with ``WO'') and 61 from USPTO (United States Patent and Trademark Office, prefixed with ``US''). 
The publication years of these patents, as illustrated in Figure~\ref{fig:statistics_1}, span the past 20 years, with over 90\% being recent patents published within the last decade.

\begin{figure}[h]
    \centering
    \includegraphics[width=0.8\linewidth]{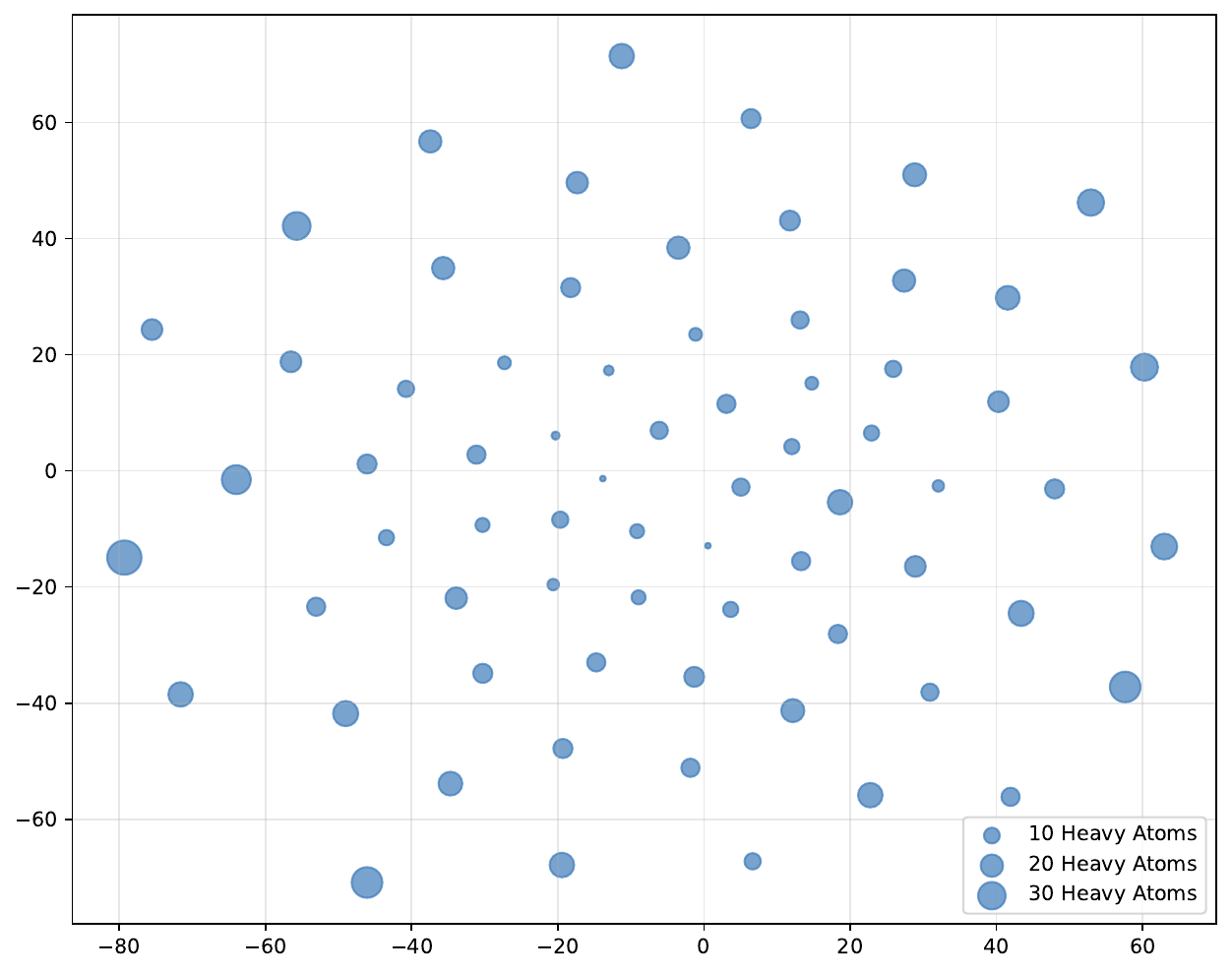}
    \caption{t-SNE Visualization of Morgan Fingerprints for Core Markush Structures in MolPatent-240.}
    \label{fig:statistics_2}
\end{figure}

To enhance the diversity of MolPatent-240, we deliberately curated patents containing a wide array of Markush structures. 
For visualization purposes, the core Markush skeletons from these 70 patents were extracted, and their Morgan fingerprints computed. 
The t-SNE dimensionality reduction of these fingerprints is presented in Figure~\ref{fig:statistics_2}, where each point represents the core Markush structure of a patent. 
The size of each point corresponds to the number of heavy atoms in the Markush structure. 
This visualization highlights substantial variation in the features of the claimed Markush skeletons, with a relatively uniform distribution in terms of both size and structural characteristics. 
The diversity of the MolPatent-240 benchmark ensures a robust and comprehensive evaluation of different algorithms.

\newpage
\section{Fine-Grained Case Study}

In Fig.~\ref{fig:main_case_stduy}, we present a case study illustrating the analytical process of PatentFinder and an example of its generated infringement report. 
For brevity, the displayed infringement report has been manually edited to fit within the layout constraints. 
In this section, we provide the model’s raw output, which includes more detailed information, such as intermediate outputs from each intelligent agent and the final summarized patent infringement report.

The key information from the patent, as extracted by the \textbf{Sketch Extractor}, is as follows:

\begin{framed}
\scriptsize
\begin{lstlisting}
Based on the provided Patent PDF Blocks, the sections that discuss the Markush structure and its detailed claim requirements begin at index **60** and end at index **88**.

These indices correspond to the detailed description of the molecular structure, variations, substitutions, and specific conditions related to the Markush groups in the patent.

The core Markush structure claimed in the patent is represented by the following SMILES notation with specified R-groups:
`**N(**C1(*****1)*)*<sep><a>0:B[5]</a><a>1:B[4]</a><a>3:B[2]</a><a>4:B[1]</a><a>6:A[1]</a><a>7:A[2]</a><a>8:A[3]</a><a>9:A[4]</a><a>10:A[5]</a><a>11:D[1]</a><a>12:B[3]</a>`

According to the google patent database, the core Markush structure claimed in the patent is represented by the following SMILES notation with specified R-groups:
`*CN(*)CCC1(*)CC(*)(*)OC2(CCCC2)C1<sep><a>0:B[5]</a><a>3:B[3]</a><a>7:D[1]</a><a>10:R[21]</a><a>11:R[22]</a>`

Below are the claim requirements discussed in the google patent database:
```markdown

Claims (36)
-----------

What is claimed is:
 
1. A compound having a formula of

\begin{molecule}
\caption{C1(*)(*)OC2(CCCC2)CC(*)(CCN(C*)*)C1<sep><a>1:R[21]</a><a>2:R[22]</a><a>11:D[1]</a><a>16:B</a><a>17:B</a>}
\end{molecule}

or a pharmaceutically acceptable salt thereof, wherein:
R21 and R22 are independently H or CH3;
D1 is an optionally substituted aryl;
B3 is H or optionally substituted alkyl; and
B5 is an optionally substituted thiophenyl,
wherein a hydrogen is replaced with a deuterium.

2. The compound of claim 1, or a pharmaceutically acceptable salt thereof, wherein D1 is an optionally substituted phenyl or an optionally substituted pyridyl.

3. The compound of claim 1, or a pharmaceutically acceptable salt thereof, wherein D1 is pyridyl.

4. The compound of claim 1, or a pharmaceutically acceptable salt thereof, wherein D1 is 2-pyridyl.

5. The compound of claim 1, or a pharmaceutically acceptable salt thereof, wherein B5 is an optionally substituted thiophenyl selected from the group consisting of

\begin{molecule}
\caption{*c1cccs1}
\end{molecule}

and

\begin{molecule}
\caption{*c1ccsc1}
\end{molecule}

6. The compound of claim 1, or a pharmaceutically acceptable salt thereof, wherein B5 is

\begin{molecule}
\caption{*C1SC=CC=1<sep><a>0:<dum></a><r>0:R[23]</r><r>0:R[30]</r>}
\end{molecule}

or

\begin{molecule}
\caption{C1=CSC=C(*)C=1<sep><r>0:R[24]</r><r>0:R[30]</r><a>5:<dum></a>}
\end{molecule}

wherein R23, R24, and R30 are each independently H, OH, cycle, aryl, branched or unbranched alkyl alcohol, halo, branched or unbranched alkyl, amide, cyano, alkoxy, haloalkyl, aklylsulfonyl, nitrite, alkylsulfanyl; or R23 and R24 together form an aryl or cycle that is attached to one or more of the atoms of B5.
```
\end{lstlisting}
\end{framed}

The substructure match between the Markush claim and the query molecule, as identified by the \textbf{Substituents Matcher}, is as follows:

\begin{framed}
\scriptsize
\begin{lstlisting}
To verify the correctness of the substructure match between the Markush claim and the query molecule, we'll analyze each R-group definition and compare it with the query molecule.

**Markush Structure**:
- **SMILES**: `*CN(*)CCC1(*)CC(*)(*)OC2(CCCC2)C1`
- **Extensions**:
  - `<a>0:B[5]</a>`: Atom index 0 corresponds to **B[5]**
  - `<a>3:B[3]</a>`: Atom index 3 corresponds to **B[3]**
  - `<a>7:D[1]</a>`: Atom index 7 corresponds to **D[1]**
  - `<a>10:R[21]</a>`: Atom index 10 corresponds to **R[21]**
  - `<a>11:R[22]</a>`: Atom index 11 corresponds to **R[22]**

**Query Molecule**:
- **SMILES**: `c1ccc([C@]2(CCNCc3ccnnc3)CCOC3(CCCC3)C2)nc1`

**R-Group Mapping Extracted by Chemical Software**:
```json
{
  "B5": "c1ccnnc1",
  "B3": "[H][H]",
  "D1": "c1ccccn1",
  "R21": "[H][H]",
  "R22": "[H][H]"
}
```

**Analysis of Each R-Group**:

1. **B[5]** (`<a>0:B[5]</a>`, Atom Index 0):
   - **Markush Position**: The attachment point at atom index 0.
   - **Proposed Substituent**: `c1ccnnc1` (Pyridazine ring).
   - **Query Molecule Correspondence**:
     - In the query molecule, the fragment `c3ccnnc3` represents a pyridazine ring.
     - This ring is connected via a methylene bridge (`CCNC`) to the rest of the molecule.
     - The substitution of B[5] with `c1ccnnc1` matches this fragment.
   - **Conclusion**: The B[5] substituent is correctly assigned.

2. **B[3]** (`<a>3:B[3]</a>`, Atom Index 3):
   - **Markush Position**: The attachment point at atom index 3.
   - **Proposed Substituent**: `[H][H]` (Hydrogen).
   - **Query Molecule Correspondence**:
     - At this position, the query molecule does not have any substituent, implying it's a hydrogen atom.
   - **Conclusion**: The B[3] substituent is correctly assigned as hydrogen.

3. **D[1]** (`<a>7:D[1]</a>`, Atom Index 7):
   - **Markush Position**: The attachment point at atom index 7.
   - **Proposed Substituent**: `c1ccccn1` (Pyridine ring).
   - **Query Molecule Correspondence**:
     - The query molecule contains a pyridine ring at the outermost part of its structure, represented by `c1ccc(... )nc1`.
     - This pyridine ring is connected to the central spiro structure through the chiral carbon `[C@]`.
   - **Conclusion**: The D[1] substituent is correctly assigned.

4. **R[21]** (`<a>10:R[21]</a>`, Atom Index 10):
   - **Markush Position**: The attachment point at atom index 10.
   - **Proposed Substituent**: `[H][H]` (Hydrogen).
   - **Query Molecule Correspondence**:
     - The query molecule does not have any substituent at this position, indicating it's a hydrogen.
   - **Conclusion**: The R[21] substituent is correctly assigned as hydrogen.

5. **R[22]** (`<a>11:R[22]</a>`, Atom Index 11):
   - **Markush Position**: The attachment point at atom index 11.
   - **Proposed Substituent**: `[H][H]` (Hydrogen).
   - **Query Molecule Correspondence**:
     - Similarly, this position in the query molecule is unsubstituted, corresponding to a hydrogen atom.
   - **Conclusion**: The R[22] substituent is correctly assigned as hydrogen.

**Comparison with Neural Network Model's R-Group Mapping**:
- The neural network model provides mappings with incorrect or invalid SMILES strings for the R-groups.
- For example:
  - `"B5": "CC1C=NN=CC=1"` is an invalid SMILES for a pyridazine ring.
  - `"D1": "C1N=CC=C(C)C=1"` suggests a methyl-substituted pyridine, which doesn't match the query molecule.
- Therefore, the neural network's mappings are inaccurate.

**Conclusion**:
- The R-group mapping extracted by the chemical software accurately reflects the substituents in the query molecule.
- By substituting the R-groups into the Markush structure, we obtain the query molecule.
- All R-group values are correct and consistent with the molecule's structure.

**Verified R-Group Mapping**:
```json
{
    "B5": "c1ccnnc1",
    "B3": "[H][H]",
    "D1": "c1ccccn1",
    "R21": "[H][H]",
    "R22": "[H][H]"
}
```
\end{lstlisting}
\end{framed}

The detailed examination of the patent’s requirements, as conducted by the \textbf{Requirements Examinator}, is as follows:

\begin{framed}
\scriptsize
\begin{lstlisting}
**Analysis:**

Let's examine each R-group substitution in the query molecule and compare them with the definitions provided in the claim requirement text.

1. **B5 (Atom Index 0):**  
   - **Substituent in Query Molecule:** Pyridazinyl (`c1ccnnc1`)
   - **Definition in Claims:**
     - Claim 1 defines B5 as "an optionally substituted **thiophenyl**, wherein a hydrogen is replaced with a deuterium."
     - Claims 5-12 further specify B5 as specific thiophenyl derivatives.
   - **Assessment:**
     - Pyridazinyl is a six-membered aromatic ring containing two adjacent nitrogen atoms.
     - Thiophenyl refers to a five-membered aromatic ring containing a sulfur atom (thiophene) attached to a phenyl group.
     - The pyridazinyl group is structurally and functionally different from the thiophenyl group specified in the claims.
     - There is no indication in any of the claims that B5 can be a pyridazinyl or any heteroaryl other than thiophenyl derivatives.
   - **Conclusion for B5:** **Does not align with the claim requirements.**

2. **B3 (Atom Index 3):**  
   - **Substituent in Query Molecule:** Hydrogen (`[H][H]`)
   - **Definition in Claims:**
     - Claim 1 defines B3 as "H or optionally substituted alkyl."
   - **Assessment:**
     - The substituent is hydrogen.
   - **Conclusion for B3:** **Aligns with the claim requirements.**

3. **D1 (Atom Index 7):**  
   - **Substituent in Query Molecule:** Pyridyl (`c1ccccn1`)
   - **Definition in Claims:**
     - Claim 1 defines D1 as "an optionally substituted aryl."
     - Claims 2-4 specify D1 as "optionally substituted phenyl or optionally substituted pyridyl," with Claim 4 specifically mentioning "2-pyridyl."
   - **Assessment:**
     - The substituent is pyridyl, specifically 2-pyridyl.
   - **Conclusion for D1:** **Aligns with the claim requirements.**

4. **R21 and R22 (Atom Indices 10 and 11):**  
   - **Substituents in Query Molecule:** Both are Hydrogen (`[H][H]`)
   - **Definition in Claims:**
     - Claim 1 defines R21 and R22 as "independently H or CH3."
   - **Assessment:**
     - Both substituents are hydrogen.
   - **Conclusion for R21 and R22:** **Align with the claim requirements.**

**Overall Assessment:**

- While the substituents for B3, D1, R21, and R22 in the query molecule align with the definitions provided in the claims, the substituent for B5 does not.
- The claims specifically define B5 as an optionally substituted **thiophenyl**, and there are no provisions or dependent claims that expand this definition to include a pyridazinyl or other heteroaryl groups.
- The query molecule's B5 substituent of pyridazinyl is outside the scope of what is claimed in the patent.

**Prediction:**

Based on the analysis, the query molecule **is not covered** under the protection scope of the patent because the B5 substituent (pyridazinyl) does not meet the specific definitions provided for B5 in the claims.
\end{lstlisting}
\end{framed}

The analysis of the patent PDF block indices that substantiate the above reasoning, as provided by the \textbf{Fact Checker}, is as follows:

\begin{framed}
\scriptsize
\begin{lstlisting}
**Requirement Indices and Details**

- **Requirement Indices**: [63, 64, 69, 70, 72, 73]
- **Requirement Details**:

1. **Block Index 63**:

   *Explanation*: This block presents the general Markush structure (Formula I) claimed in the patent. It defines the core scaffold of the compounds, including the positions of variable groups such as A1-A5, B1-B5, and D1. This structure is essential for determining whether the target molecule falls within the scope of the patent.

2. **Block Index 64**:

   *Explanation*: This block provides detailed definitions for the variables A1 to A5 in the Markush structure. These definitions specify the allowed atoms or groups that can occupy each position in the core scaffold. Understanding these substitutions is crucial for comparing the target molecule's corresponding positions to the claimed compounds.

3. **Block Index 69**:

   *Explanation*: This block defines the variables B1 to B5, including B5. It states that B5 can be "alkyl, branched alkyl, halogenated alkyl, carbocycle-substituted alkyl, aryl, carbocycle, or arylalkyl." This definition is directly relevant because the B5 group in the target molecule is a pyridazine ring, which needs to be assessed against the allowed substituents for B5.

4. **Block Index 70**:

   *Explanation*: This block elaborates on the definitions of "aryl," "carbocycle," and "heterocycle," including allowable substitutions. It mentions that aryl groups can be substituted with small groups and may include heteroatoms such as nitrogen. This is pertinent to determining whether heteroaryl groups like pyridazine (present in the target molecule) are encompassed within the definitions of B5 and D1.

5. **Block Index 72**:

   *Explanation*: This block specifies that D1 is "an aryl group or a carbocycle." Since the D1 group in the target molecule is a pyridine ring (a heteroaryl group), this definition helps to confirm whether the target molecule's D1 substitution aligns with the claimed compounds in the patent.

6. **Block Index 73**:

   *Explanation*: This block provides examples and clarifications for the term "aryl group," noting that it can be a monocyclic or bicyclic aromatic group that may contain heteroatoms (heteroaryl). It includes structures representing various heteroaryl groups, which directly relates to the pyridazine and pyridine rings in the target molecule, impacting the assessment of infringement.

**Summary of Relevance to the Infringement Analysis**

The blocks identified above contain the key definitions and descriptions necessary to determine whether the target molecule is covered by the patent claims. Specifically:

- **B5 Group Analysis**: Blocks 69, 70, and 73 are crucial for assessing the B5 substitution. The patent defines B5 as potentially being an "aryl" group, and further explanations include heteroaryl groups like pyridazine. Thus, the pyridazine ring in the target molecule may fall within the scope of B5 as defined in the patent.

- **D1 Group Analysis**: Blocks 72 and 73 clarify that D1 can be an aryl group, including heteroaryl groups such as pyridine. This confirms that the pyridine ring in the target molecule aligns with the D1 definitions in the patent.

- **Overall Structure Alignment**: Blocks 63 and 64 establish the core structure and positions of substitutions, allowing for a direct comparison between the patent claims and the target molecule's structure.

**Conclusion**

Based on the definitions provided in the identified blocks, the target molecule appears to be encompassed by the Markush structure claimed in the patent. The allowances for heteroaryl groups in positions B5 and D1 suggest that the substitutions present in the target molecule are within the scope of the patent claims.

---

**Final Notes**

- The extracted blocks focus on the detailed claim requirements specifying the allowed variations and substitutions for the Markush groups, directly impacting the infringement analysis.

- The blocks are sorted by their contribution to the analysis, starting with the core structure and moving through the specific group definitions.

- Each requirement detail corresponds precisely to the block index, providing a clear and concise explanation of its relevance.
\end{lstlisting}
\end{framed}
\newpage

The final infringement report, summarized by the \textbf{Infringement Reporter}, is as follows:

\begin{framed}
\scriptsize
\begin{lstlisting}
**Infringement Report**

---

**1. Structure of the Query Molecule**

The query molecule has the following structure:

- **SMILES**: `c1ccc([C@]2(CCNCc3ccnnc3)CCOC3(CCCC3)C2)nc1`
- **IUPAC Name**: N-(pyridazin-4-ylmethyl)-2-[(9R)-9-pyridin-2-yl-6-oxaspiro[4.5]decan-9-yl]ethanamine

This structure consists of:

- A spirocyclic core with an oxaspiro[4.5]decane ring system.
- A 2-pyridyl group attached to the spiro center.
- An ethylamine chain connecting to a pyridazin-4-ylmethyl group.

---

**2. Structure of the Core Markush Protected in the Patent and Corresponding R-group Definitions**

The core Markush structure claimed in the patent is represented by the following SMILES notation with specified R-groups:

- **Markush SMILES**: `*CN(*)CCC1(*)CC(*)(*)OC2(CCCC2)C1`
- **R-group Definitions**:

  | Atom Index | R-group | Description         |
  |------------|---------|---------------------|
  | 0          | B5      | B[5] substituent    |
  | 3          | B3      | B[3] substituent    |
  | 7          | D1      | D[1] substituent    |
  | 10         | R21     | R[21] substituent   |
  | 11         | R22     | R[22] substituent   |

**R-group Definitions from Patent Claims:**

- **B5**: An optionally substituted thiophenyl, wherein a hydrogen is replaced with a deuterium.
- **B3**: Hydrogen (H) or an optionally substituted alkyl.
- **D1**: An optionally substituted aryl.
- **R21** and **R22**: Independently hydrogen (H) or methyl (CH3).

---

**3. Values of Each R-group in the Query Molecule**

Based on the substructure matching, the R-groups in the query molecule are as follows:

- **B5** (Atom Index 0):
  - **Substituent**: Pyridazinyl (`c1ccnnc1`)
- **B3** (Atom Index 3):
  - **Substituent**: Hydrogen (`[H]`)
- **D1** (Atom Index 7):
  - **Substituent**: 2-Pyridyl (`c1ccccn1`)
- **R21** (Atom Index 10):
  - **Substituent**: Hydrogen (`[H]`)
- **R22** (Atom Index 11):
  - **Substituent**: Hydrogen (`[H]`)

---

**4. Alignment of Substitutions with Patent Coverage Conditions**

**a. B5 Substituent**

- **Query Molecule Substituent**: Pyridazinyl (`c1ccnnc1`)
- **Patent Definition**: An optionally substituted **thiophenyl**, wherein a hydrogen is replaced with a deuterium.
- **Analysis**:
  - **Pyridazinyl** is a six-membered aromatic ring containing two adjacent nitrogen atoms.
  - **Thiophenyl** refers to a five-membered aromatic ring containing a sulfur atom (**thiophene**) attached to a phenyl group.
  - The pyridazinyl group is structurally different from the thiophenyl group specified in the patent.
  - The patent claims do not extend B5 to include pyridazinyl or other heteroaryl groups besides thiophenyl derivatives.
- **Conclusion for B5**: The B5 substituent **does not align** with the patent's claim requirements.

**b. B3 Substituent**

- **Query Molecule Substituent**: Hydrogen (`[H]`)
- **Patent Definition**: Hydrogen (H) or an optionally substituted alkyl.
- **Analysis**:
  - The substituent in the query molecule is hydrogen.
- **Conclusion for B3**: The B3 substituent **aligns** with the patent's claim requirements.

**c. D1 Substituent**

- **Query Molecule Substituent**: 2-Pyridyl (`c1ccccn1`)
- **Patent Definition**: An optionally substituted aryl, with dependent claims specifying optionally substituted pyridyl, particularly **2-pyridyl**.
- **Analysis**:
  - The substituent matches the 2-pyridyl group specified in Claim 4.
- **Conclusion for D1**: The D1 substituent **aligns** with the patent's claim requirements.

**d. R21 and R22 Substituents**

- **Query Molecule Substituents**: Both are Hydrogen (`[H]`)
- **Patent Definition**: Independently hydrogen (H) or methyl (CH3).
- **Analysis**:
  - Both substituents are hydrogen, fitting within the specified options.
- **Conclusion for R21 and R22**: Both substituents **align** with the patent's claim requirements.

---

**5. Other Patent Requirements**

**Deuterium Substitution Requirement**

- **Patent Claim**: In Claim 1, it is specified that "a hydrogen is replaced with a deuterium."
- **Analysis**:
  - The query molecule does **not** contain any deuterium atoms; all hydrogens remain unmodified.
  - The requirement specifies that at least one hydrogen atom in the compound must be replaced with a deuterium atom.
- **Conclusion**: The query molecule **does not** meet the deuterium substitution requirement of the patent.

---

**6. Definitive Conclusion**

After a comprehensive analysis of the query molecule against the patent's claims:

- **B5 Substituent Misalignment**: The query molecule's B5 substituent is a pyridazinyl group, which does not align with the patent's requirement of an optionally substituted thiophenyl.

- **Deuterium Substitution Requirement Not Met**: The query molecule lacks the necessary deuterium substitution specified in the patent claims.

**Final Verdict**:

The query molecule **is not covered** under the protection scope of the patent due to the failure to meet essential claim requirements regarding the B5 substituent and the mandatory deuterium substitution.

---

**Recommendation**:

Based on the above analysis, it is concluded that the query molecule does not infringe upon the patent in question. It falls outside the scope of the patent claims and does not meet the specific structural and compositional requirements stipulated.
\end{lstlisting}
\end{framed}

\section{Reproducibility}



\begin{figure}[h]
    \centering
    \includegraphics[width=\linewidth]{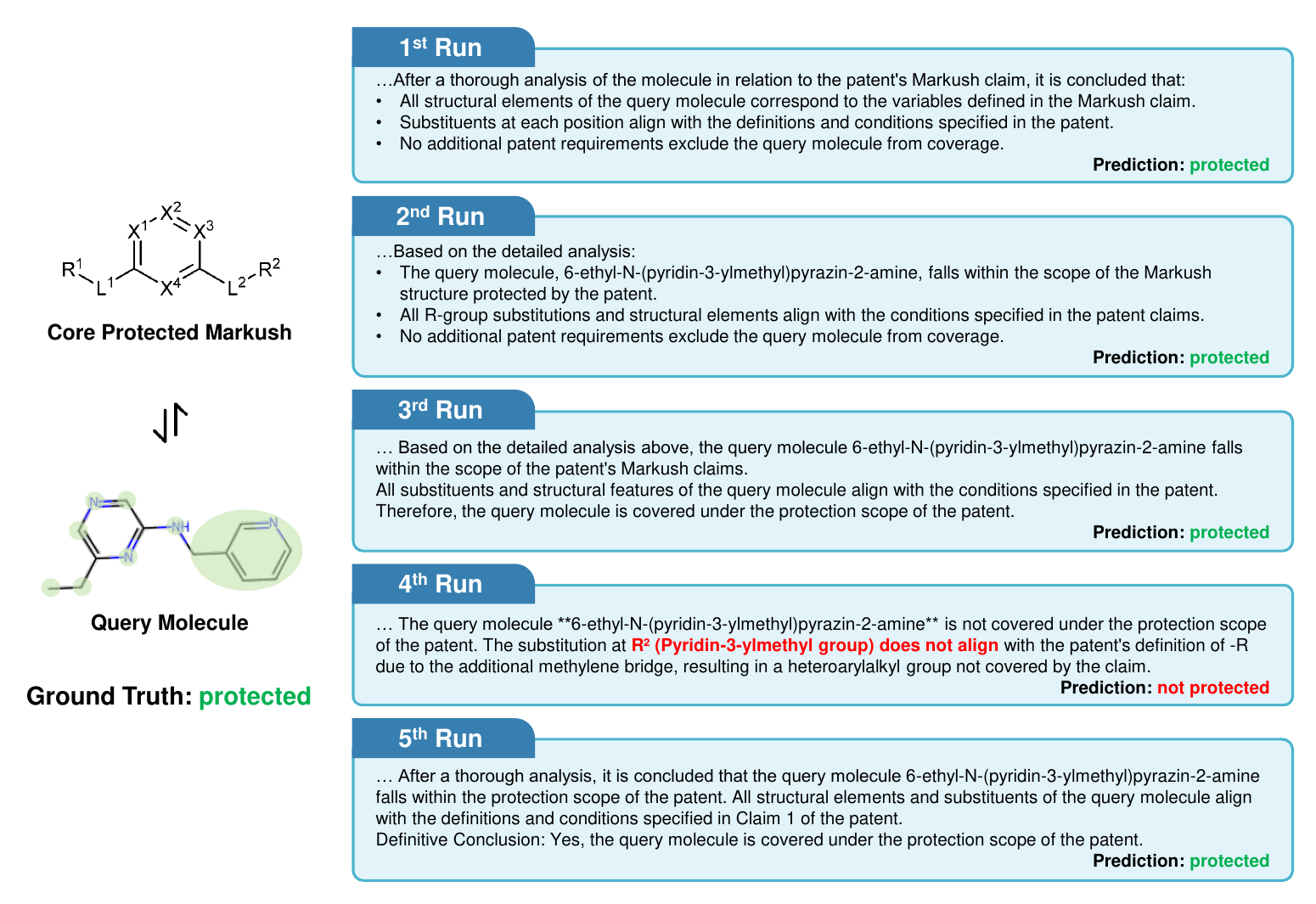}
    \caption{Various patent infringement reports generated by PatentFinder for the same sample. For clarity, only the concluding paragraphs from each report are presented.}
    \label{fig:reproducibility}
\end{figure}

When applying multi-agent systems to scientific discovery tasks, a crucial aspect is evaluating the reproducibility of their outputs to ensure stability in real-world applications. 
To assess the reproducibility of PatentFinder, we randomly select a sample from the MolPatent-240 dataset and conducted five independent patent infringement analyses.
As illustrated in Figure~\ref{fig:reproducibility}, all substituents comply with the definitions provided in the corresponding patent and are therefore protected under its claims. 
The primary challenges in this case include:
(1) The Markush structure predominantly contains undefined substituents, posing significant challenges for structural recognition and match prediction by the tool model.
(2) The patent specifies distinct definitions for the $L^1R^1$ and $L^2R^2$ sides, resulting in special cases mentioned in Figure~\ref{fig:markush_matcher_exp}f and ~\ref{fig:markush_matcher_exp}g within the Markush structure. 
These require the language model to integrate patent content with substituent-matching reasoning.

For the experiments, we employed OpenAI-o1 as the backbone language model of PatentFinder, with a temperature of $1.0$. 
Figure~\ref{fig:reproducibility} illustrates the five patent infringement reports generated during the analysis. 
The results show that in 4 out of 5 cases, PatentFinder successfully matched the molecular structure with the patent’s claim requirements, arriving at the correct conclusion. 
However, in the fourth report, an error occurred when PatentFinder misjudged the $R^2$ substituent as non-compliant with the requirement of being an ``optionally substituted 4- to 6-membered heterocyclic ring having 1-2 heteroatoms''.
This issue arises from limitations in the language model’s understanding of molecular structures. 
Consequently, the performance of PatentFinder can be further enhanced through advancements in the backbone language model and by adapting scientific language models. 
Additionally, a simple majority voting strategy can improve the stability of PatentFinder. 
By generating multiple infringement reports and deciding the final prediction based on majority votes, the system can achieve greater reliability. 
For instance, in this case, PatentFinder could correctly predict the molecule as being protected by the patent in 4 out of 5 analyses, yielding the correct outcome.

\section{Bad Case Analysis}

\begin{figure}[h]
    \centering
    \includegraphics[width=\linewidth]{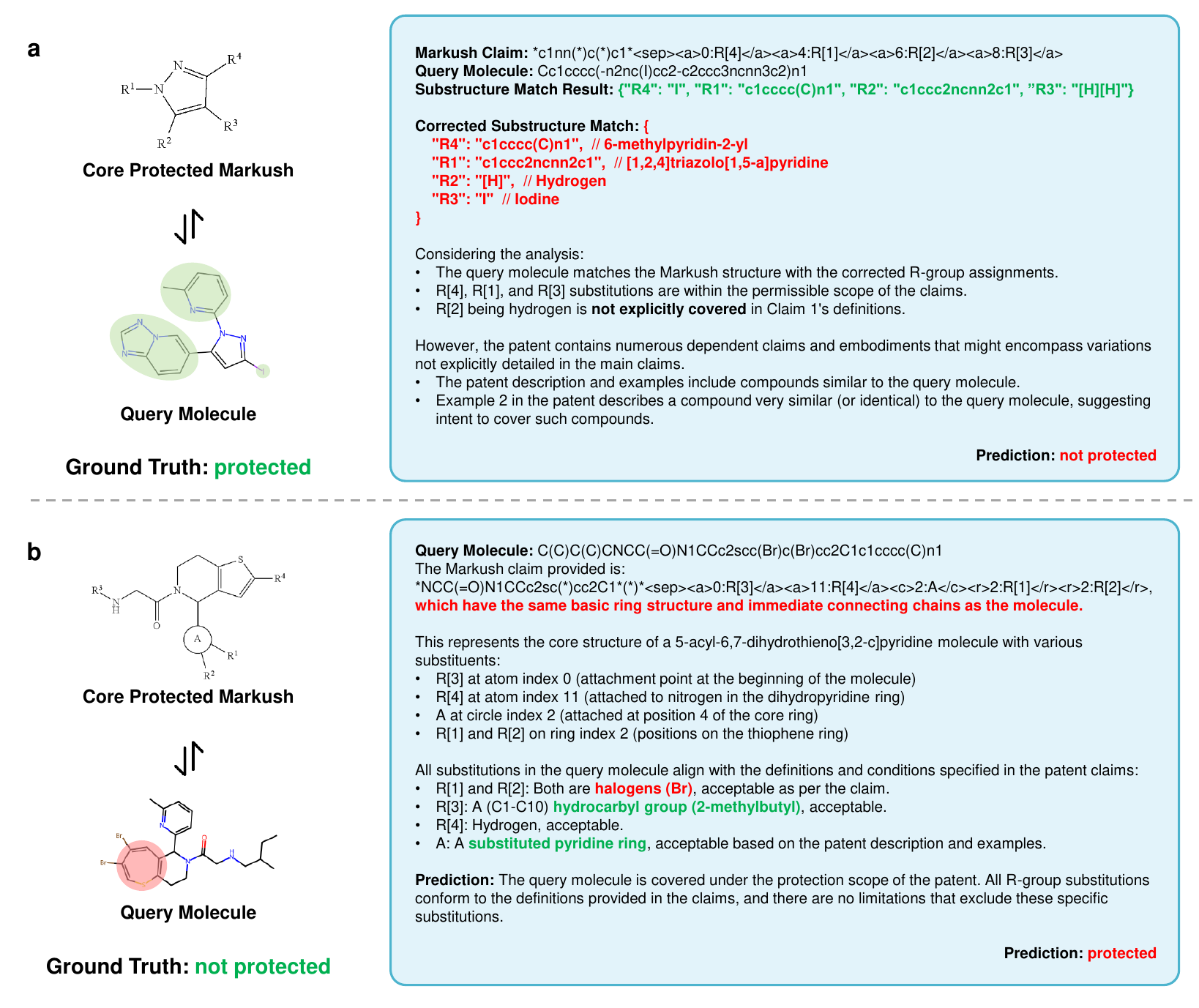}
    \caption{Examples of incorrect predictions by PatentFinder on the MolPatent-240 dataset. a) The Substituents Matcher erroneously alters the correct prediction by MarkushMatcher, resulting in a false negative prediction; b) The Requirements Examinator fails to detect the mismatch between the query molecule and the skeleton of the Markush structure, leading to a false positive prediction. For clarity, both the infringement reports depicted are redacted.}
    \label{fig:bad_case}
\end{figure}

Experiments on the MolPatent-240 dataset demonstrate that while PatentFinder surpasses baseline models in overall accuracy for patent infringement prediction, it is not yet capable of perfectly resolving all molecular infringement cases. 
To better understand the limitations of PatentFinder and guide future research, we analyze two failed predictions from MolPatent-240, as illustrated in Figure~\ref{fig:bad_case}.

\textbf{Sample a.}
The molecular structure in this example perfectly matches the Markush scaffold, and its substituents meet all patent requirements. 
Therefore, it is supposed to be classified as protected. 
During PatentFinder’s reasoning process, the MarkushMatcher module correctly identifies the substituent matches. 
However, the Substituents Matcher module erroneously modifies the match results during the correction step, ultimately misclassifying the molecule as not protected. 
This reveals a tendency of the language model to overconfidently override correct predictions provided by specialized submodules.

\textbf{Sample b.}
This molecule contains a sulfur-containing heterocyclic ring (highlighted in red area) that does not match the corresponding structure in the Markush claim, and it should therefore be classified as not protected. 
However, the Requirements Examinator module incorrectly concluded that the query molecule’s scaffold matches the Markush structure. 
This misjudgment led to subsequent errors in substituent matching, resulting in a false positive prediction that the molecule is protected. 
This highlights the model’s limitations in analyzing complex chemical structures.

Analysis of these failure cases reveals that while PatentFinder can successfully conduct parts of its reasoning process and remains faithful to the patent’s original requirements, certain steps still lead to errors. 
Despite the incorrect conclusions, the intermediate reasoning steps can provide valuable insights for drug discovery researchers, showcasing the high interpretability and potential utility of the PatentFinder framework.

\section{Limitations}


While PatentFinder demonstrates promising performance in patent infringement analysis, surpassing traditional algorithms and baseline methods relying solely on text-based language models, several challenges remain for its deployment in drug discovery workflows:
(1) \textbf{Instability Introduced by LLMs:}
Despite the multi-agent reasoning system and the incorporation of domain-specific tool models designed to mitigate hallucinations and enhance factual accuracy, LLMs can still produce responses with logical flaws or factual inaccuracies. 
Neither the multi-agent system nor the tool models can entirely eliminate the possibility of errors during the LLM’s reasoning process, posing a risk to the robustness of the framework.
(2) \textbf{Difficulty in Evaluating the Analytical Process:}
Patent infringement assessment involves complex reasoning, and evaluating each intermediate step of the multi-agent system is challenging. 
Although this paper introduces the MolPatent-240 dataset to assess model performance, the evaluation is limited to treating patent infringement determination as a binary classification task. 
Due to the inherent complexity of patent analysis, it is difficult to establish ground truth for each intermediate step and design appropriate evaluation metrics. 
This limitation makes further assessment of PatentFinder highly dependent on manual labor, constraining the speed and scale of evaluations.
(3) \textbf{Dependence on Upstream Patent Retrieval Systems:}
PatentFinder requires relevant patents to be gathered by an upstream patent retrieval system for comparison. 
Errors in this retrieval process -- such as failing to provide patents pertinent to the target molecule -- may result in incorrect conclusions, such as classifying a protected molecule as unprotected. 
Such errors could lead to significant legal risks.

Despite its advancements, PatentFinder cannot fully replace human effort in this domain. 
However, with ongoing progress in artificial intelligence, more powerful backbone language models and advanced molecular analysis tools are expected to enhance PatentFinder’s capabilities in patent analysis and molecular understanding. 
These advancements will further unlock the potential of its multi-agent architecture, improving both performance and reliability in patent infringement analysis and related applications.

\end{document}